\documentclass[conference]{IEEEtran}
\IEEEoverridecommandlockouts
\usepackage{amsmath,amssymb}
\usepackage[T1]{fontenc}
\usepackage[utf8]{inputenc}
\usepackage{latexsym}
\usepackage{soul}
\usepackage{tikz}
\usepackage{xfrac}
\usepackage{xcolor}
\usepackage{hyperref}
\usepackage{lipsum}
\usepackage{layouts}
\usepackage{mathtools}
\usepackage{diagbox}
\usepackage{xspace}
\usepackage{makecell}
\newcommand{\coverage}{\emph{coverage}\xspace}
\newcommand{\Coverage}{\emph{Coverage}\xspace}
\newcommand{\precision}{\emph{precision}\xspace}
\newcommand{\Precision}{\emph{Precision}\xspace}
\newcommand{\reliability}{\emph{reliability}\xspace}

\newcommand{\seconds}[1]{\mbox{#1\,\textrm{s}}}
\newcommand{\vx}{\mathbf{x}}
\newcommand{\paren}[1]{\left(#1\right)}
\newcommand{\floor}[1]{{\lfloor #1 \rfloor}}
\newcommand{\ceil}[1]{{\lceil #1 \rceil}}
\newcommand{\op}[1]{\operatorname{#1}}
\newcommand{\opp}[2]{\operatorname{#1}\paren{#2}}
\newcommand{\ME}{MAP-Elites\xspace}
\newcommand{\explanation}[1]{\begin{center}\parbox{0.95\linewidth}{#1}\end{center}}

\definecolor{orange}{HTML}{FFd080}

\newcommand{\imagelabel}[1]{\makebox[0pt][r]{\raisebox{4pt}{#1\hspace{4pt}}}}

\begin{document}

\title{Multi-objective Analysis of MAP-Elites Performance}

\author{\IEEEauthorblockN{Eivind Samuelsen}
\IEEEauthorblockA{\textit{Department of Informatics} \\
\textit{University of Oslo}\\
Oslo, Norway \\
email address}
\and
\IEEEauthorblockN{Kyrre Glette}
\IEEEauthorblockA{\textit{Department of Informatics} \\
\textit{University of Oslo}\\
Oslo, Norway \\
kyrrehg@ifi.uio.no}}

\maketitle

\begin{abstract}
In certain complex optimization tasks, it becomes necessary to use multiple measures to characterize the performance of different algorithms. 
This paper presents a method that combines ordinal effect sizes with Pareto dominance to analyze such cases.  Since the method is ordinal, it can also generalize across different optimization tasks even when the performance measurements are differently scaled.  Through a case study, we show that this method can discover and quantify relations that would be difficult to deduce using a conventional measure-by-measure analysis. 
This case study applies the method to the evolution of robot controller repertoires using the MAP-Elites algorithm.
Here, we analyze the search performance across a large set of parametrizations; varying mutation size and operator type, as well as map resolution, across four different robot morphologies. 
We show that the average magnitude of mutations has a bigger effect on outcomes than their precise distributions.
\end{abstract}

\section{Introduction}
Search algorithms can be applied to a multitude of tasks in engineering, science and other fields, from scheduling and optimization tasks to design and creativity~\cite{Konak2006,Renner2003}.
While in some straightforward cases it is sufficient for the search to end up with the single best solution, or a set of nondominated solutions in the case of a multi-objective optimization task, it would in some cases be an advantage to be able to investigate a range of different and high-performing solutions from the same search process.

In this context, algorithms like Novelty Search with Local Competition\cite{Lehman2011} and \ME \cite{Mouret2015} have been proposed, which explore and keep track of solutions that are high-performing but different according to a behavior criterion.
These algorithms, named Quality Diversity (QD) or Illumination algorithms, have been successfully applied to e.g. 
avoiding deception and promoting diversity in hard search domains~\cite{Lehman2013}, for the generation of a large set of qualitatively different artworks \cite{Nguyen2015}, or for exploring the design space of airfoils~\cite{Gaier2017}.
It should be noted that the exploratory nature of these algorithms can have a dual purpose--either solely as a diversity-enhancing functionality, serving as stepping stones for the algorithm to overcome deceptive search spaces~\cite{Lehman2013}--or also for inspecting or exploiting the generated repertoire of solutions~\cite{Mouret2015}.

QD algorithms have been particularly successful in the Evolutionary Robotics (ER) field \cite{Doncieux2015}, with applications such as soft robot evolution \cite{Mouret2015}, damage recovery \cite{Cully2015}, and locomotion repertoire generation~\cite{Cully2016, Duarte2017}.
Recently, \cite{Pugh2016} compared the performances of a range of QD variants, and \cite{Cully2017} proposed a unifying framework for QD. 


\ME \cite{Mouret2015}, searches a space of user-defined features, or \emph{behaviors}, by discretizing it into a grid.
As the algorithm progresses, the cells in this grid are progressively filled with solutions according to their position in the behavior space, replacing any solution already associated with the cell only if the new solution is better according to some user-defined quality measure.
This property stands out as particularly attractive for some applications, since the result of the search is a regular grid and one can easily locate a cell with a desired behavior, e.g. for use in locomotion repertoire generation tasks~\cite{Duarte2017}.


QD algorithms such as \ME naturally give rise to multiple ways of measuring the their performance:  one would like to know how diverse the solutions are in terms of how much of the behavior space is covered, as well as the quality of the solutions. 
These aspects are covered by \cite{Mouret2015} by the \coverage and \precision measures, respectively, along with a third measure called \emph{global reliability}. \cite{Pugh2016} reduces these quality and diversity criteria down to a single measure, called QD-score, that increases with any improvement in the two areas.
%

In multi-objective optimization (MOO), it is known that any way of combining multiple objective functions into one gives challenges related to scaling and potential loss of solutions, compared to a Pareto-based treatment \cite{Konak2006}.
We argue that the same concept also applies when comparing algorithms on multiple performance measures, and therefore propose a Pareto dominance-based analysis method that takes this explicitly into consideration.

Usually MOO will deal with deterministic objective function evaluations, where one can say with absolute certainty if one solution dominates the other. Stochastic algorithms such as \ME, however, can produce very varying results, so instead of a single value for each performance measure, multiple runs of the algorithms will result in a set of different values. Using Cliff's delta \cite{Cliff1993}, an ordinal statistic closely related to the Vargha-Delaney effect size \cite{Vargha2000} suggested for this use by \cite{Neumann2015}, we are able to describe to what degree the outcomes of one algorithm dominates those of another. 

We have earlier co-evolved robot morphologies and controllers, and produced real-world instances of these \cite{Samuelsen2015}.
To fully make use of a new morphology, it would be relevant to learn a repertoire of walking behaviors, such as being able to move forwards and backwards and turn with different speeds, much in the same vein as \cite{Cully2016} and \cite{Duarte2017}.
Therefore, as a case study we employ \ME to generate controller repertoires. 
We illustrate that the proposed effect size, being an ordinal statistic, is able to generalize the results across four distinct robot morphologies.

One important parameter in \ME is the map resolution, which decides how many cells to divide the map into along each behavior axis. Lower resolution would perhaps result in more difficulty reaching new cells, but it might also reduce the number of evaluations needed to fill the map, simply because there are fewer cells to fill. 

Another aspect that could have a significant impact on the performance of the evolutionary search, is the choice of mutation operator.
Some studies in ER always apply random perturbations to every parameter, while others only apply the perturbations to each parameter with a certain probability. 
Both approaches has their merits, 
and we seek to understand how the two affect the performance of our search. 
We also test a range of values for the mutation size, i.e. the $\sigma$ of the Gaussian distribution. This is partly to control for effects that do not generalize across mutation sizes, but also to shed light on possible performance trade-offs.

The results from our experimental runs are reported in effect size tables, allowing for statistical comparisons, as well as visualizing them in parametric plots, giving a good overview of the multi-objective performance aspects.
We also visualize an excerpt of the results in a measure-by-measure fashion, to contrast with the conventional analysis.

To summarize, this paper makes two major contributions:
First, we demonstrate how we can apply a Pareto dominance based Cliff's delta calculation for comparing performances of \ME searches.
Secondly, we demonstrate the usefulness of these methods with a case study in evolutionary robotics, with a particular focus on different mutation schemes and how these affect performance.

\begin{table}
\caption{Experiment setup}
\small
\begin{tabular}{p{1.45in}p{1.35in}}
\hline
PhysX version & 3.4 \\
Ground-robot friction & 0.3 / 0.3 \\
Timestep & \seconds{$128^{-1}$} \\
\hline
Control system period & \seconds{1} \\
Pre-evaluation periods & 1 \\
Evaluation periods & 4 \\
Samples per period & 4 \\
\hline
\emph{Behavior feature} & \emph{Map extents} \\
Turn rate & $\pm3\,\mathrm{rad/s}$ \\
Adjusted forward speed & $\pm0.75\,\mathrm{m/s}$ \\ 
\hline
Performance measure & Weighed penalties for \\ & \enskip- Large body pitch \\& \enskip- Low body height \\& \enskip- Sideways movement \\
\hline
Initial population & 100 \\
Initial mutation & all-hard \\
Initial $\sigma$ & 0.5 \\
Total evaluations per run & 20000 \\
\hline
Map resolutions & 5$\times$5, 7$\times$7, 9$\times$9 \\
$\sigma$ values & 0.01, 0.1, 0.2, 0.4, 0.8 \\
Mutation types & all, some \\
\begin{tabular}{@{}p{0.9in}r@{}}
Robots & robot2 \\
       & robot3 \\
       & robot4 \\
       & robot5 \\
\end{tabular} & 
\begin{tabular}{@{}l@{}}
4 legs, 6 joints \\
4 legs, 9 joints \\
4 legs, 10 joints \\
6 legs, 14 joints \\
\end{tabular} \\
Total combinations & 240 \\
Runs per combination & 12 \\
\hline
\end{tabular}
\label{tbl:setup}
\end{table}

\section{Methods}

We implemented the MAP-Elites algorithm as described in~\cite{Mouret2015} to find repertoires of robot gaits that effect useful behaviors. We then analyze the effect of different map resolutions, mutation operators and mutation sizes on the quality of the repertoires for four different robots. The robots and the simulator are described in \autoref{method:robots}. The gait behavior and performance measures are defined in \autoref{method:features}. The mutation operators are described in \autoref{method:mutators}. The measures used to judge repertoire quality is defined in \autoref{method:meas}. Finally, the methods used to analyze the results are presented in sections \ref{method:effect-size} and \ref{method:pareto}. A summary of key details is shown in \autoref{tbl:setup}. 
The experiment data and R source code for the analysis, along with videos of evolved repertoires, is available online.\footnote{Temporary address for review: \url{https://folk.uio.no/eivinsam/data/ALIFE18}}

\subsection{Simulated Robots}\label{method:robots}
\begin{figure}
\mbox{
\includegraphics[width=1.6in]{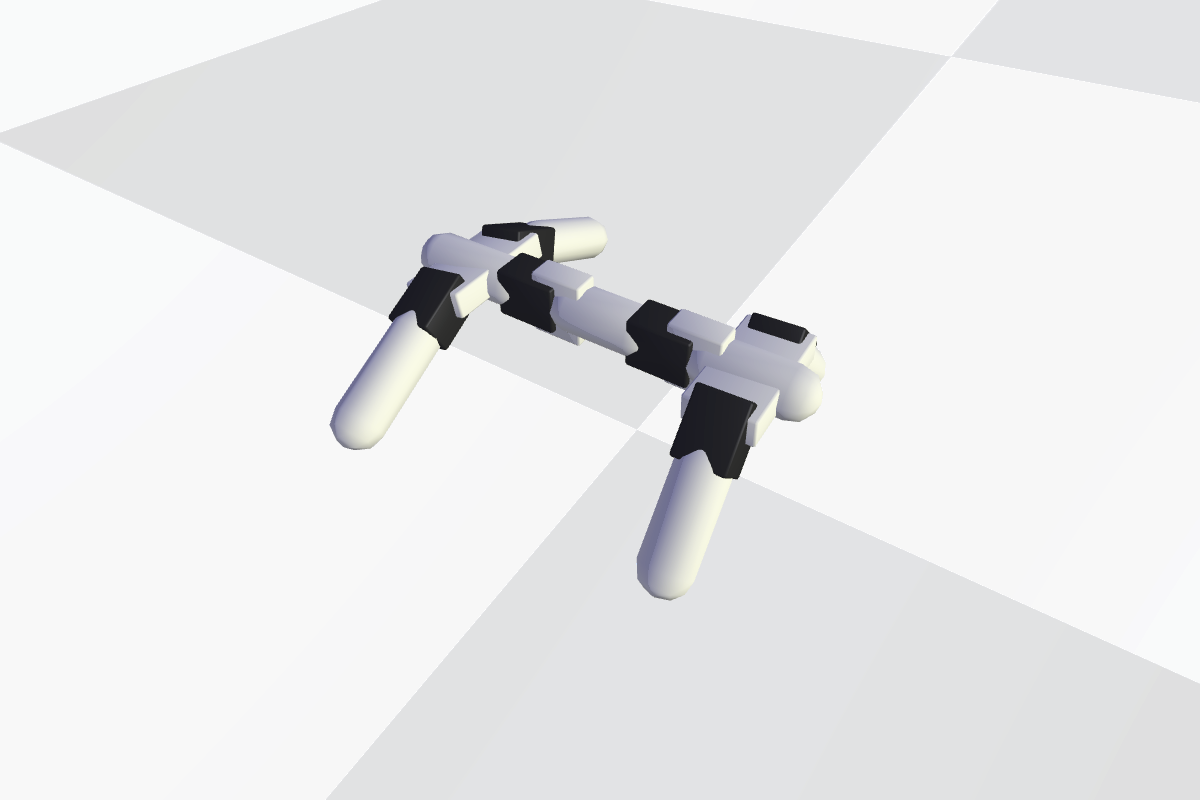}\imagelabel{robot2} 
\includegraphics[width=1.6in]{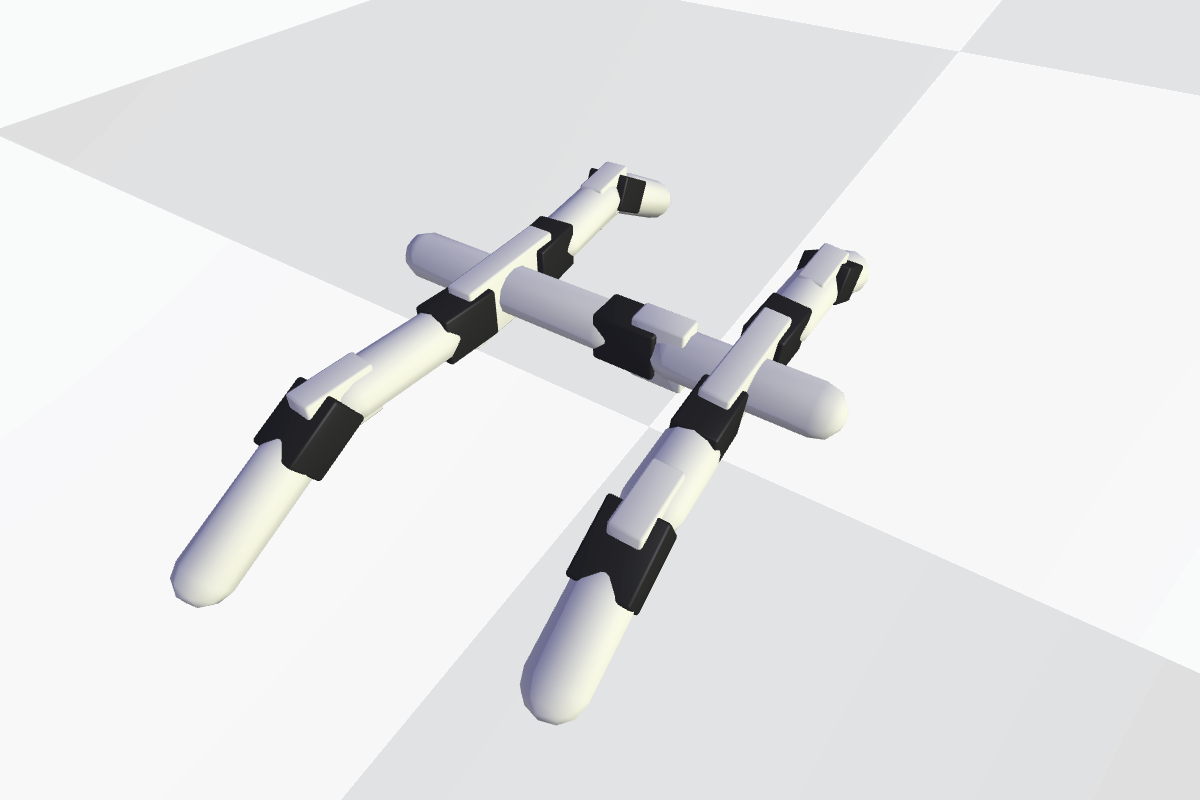}\imagelabel{robot3} \\
}
\mbox{
\includegraphics[width=1.6in]{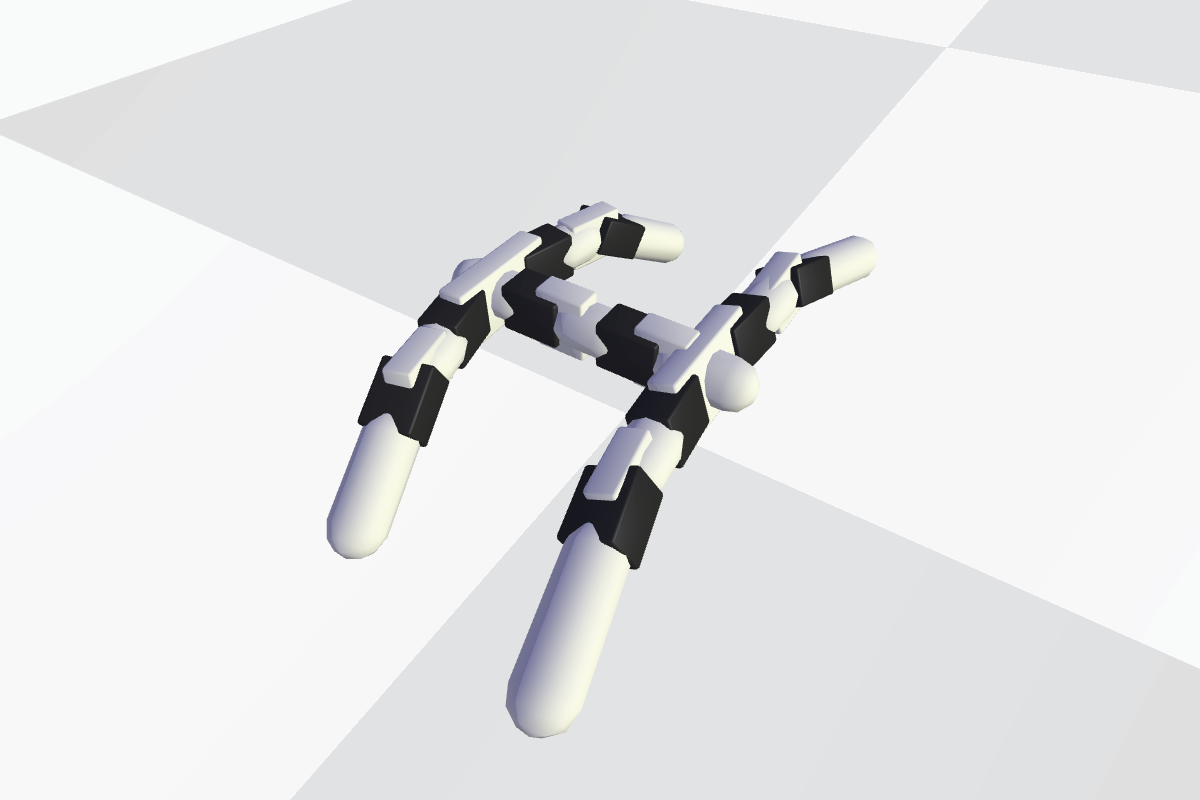}\imagelabel{robot4} 
\includegraphics[width=1.6in]{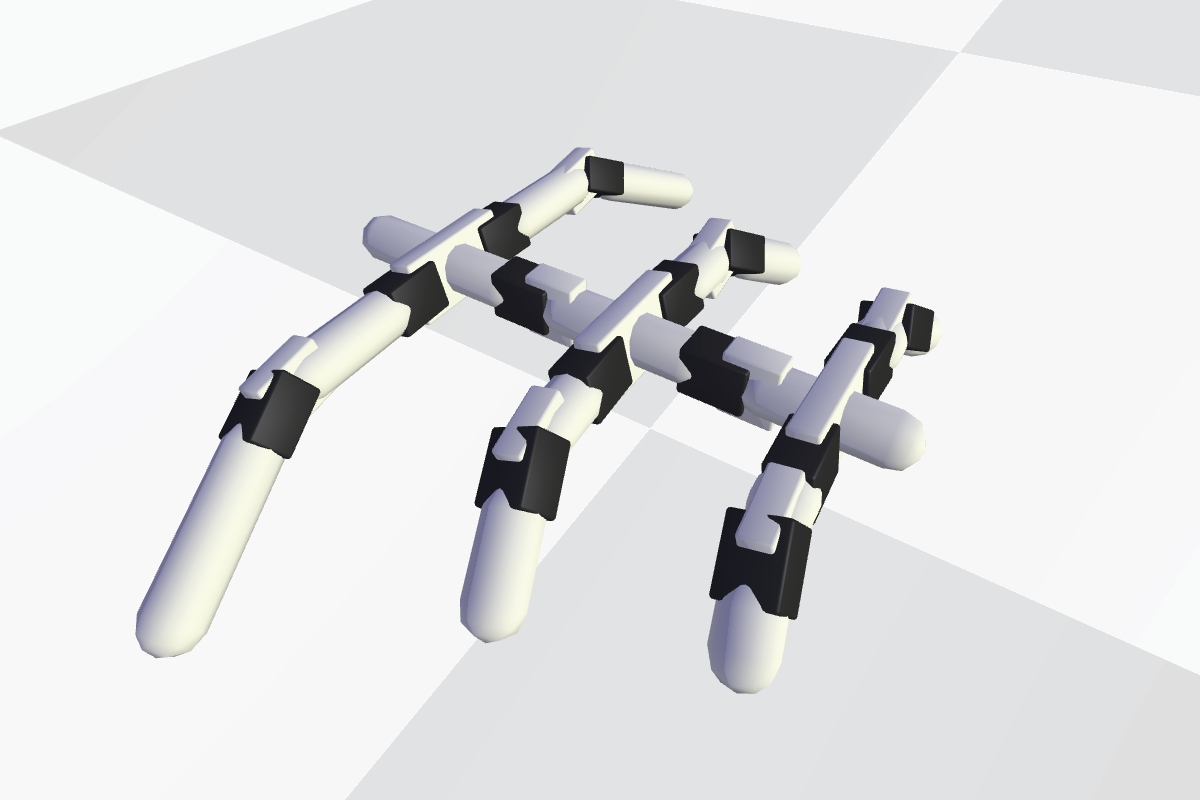}\imagelabel{robot5}
}
\caption{The four robot morphologies}
\label{fig:robots}
\end{figure}

The algorithm was run on four robot morphologies selected from \cite{Samuelsen2015}, illustrated in \autoref{fig:robots}. 
The morphologies have retained their original numbering. 
Using the PhysX simulator, the subject robot is simulated on a infinite flat plane. When given a new gait to evaluate by MAP-Elites, it first runs the gait for one period before starting to sample position and orientation. 

Sampling continues until a given number of periods have elapsed. Based on the samples, behavior and performance (\autoref{method:features}) are given to MAP-Elites. This repeats without restarting the simulator, until the run ends when a certain number of evaluations has been reached or the simulator detects that the robot has been in a tilted or flipped state for 20 consecutive evaluations. After each evaluation the current \coverage and \precision (\autoref{method:meas}) is logged for analysis.

\subsection{Control System}\label{method:control}
Each joint is controlled by an open-loop controller with four parameters: phase offset $\phi$, duty cycle $D$, and two extreme values $v_0$ and $v_1$, as shown in Figure \ref{fig:control}. The duty cycle is encoded as a continuous parameter $d$ such that 
\[ 
 D = 
 \begin{cases} 
 d-\floor{d} & \floor{d}\text{ is even} \\
 \ceil{d}-d  & \text{otherwise} \\
 \end{cases}
\]
as illustrated in Figure \ref{fig:dwrap}. This lets $d$ be mutated freely as a continuous variable while mapping it to the correct range without discontinuities and without getting stuck at the extreme values. The parameters are encoded symmetrically, so that each joint on the left side of the body share duty cycle and amplitude parameters with the corresponding joint on the right side. The phase offset is coded differentially, i.e. it is $\phi$ on the left side and $\phi + \Delta\phi$ on the right.

\begin{figure}
\begin{tikzpicture}[x=1pt,y=1pt]
\definecolor{fillColor}{RGB}{255,255,255}
\path[use as bounding box,fill=fillColor,fill opacity=0.00] (0,0) rectangle (240.90, 67.45);
\begin{scope}
\path[clip] ( 23.40, 13.51) rectangle (236.40, 62.95);
\definecolor{drawColor}{RGB}{0,0,0}

\path[draw=drawColor,line width= 0.6pt,line join=round] ( 33.08, 57.71) --
	(100.86, 15.76) --
	(125.06, 60.70) --
	(197.67, 15.76) --
	(221.88, 60.70) --
	(226.72, 57.71);
\definecolor{drawColor}{RGB}{190,190,190}

\path[draw=drawColor,line width= 0.6pt,line join=round] ( 23.40, 15.76) -- (236.40, 15.76);
\definecolor{drawColor}{RGB}{0,0,0}

\node[text=drawColor,anchor=base,inner sep=0pt, outer sep=0pt, scale=  1.10] at (210.26, 16.52) {$v_0$};
\definecolor{drawColor}{RGB}{190,190,190}

\path[draw=drawColor,line width= 0.6pt,line join=round] ( 23.40, 60.70) -- (236.40, 60.70);
\definecolor{drawColor}{RGB}{0,0,0}

\node[text=drawColor,anchor=base,inner sep=0pt, outer sep=0pt, scale=  1.10] at (210.26, 54.62) {$v_1$};
\definecolor{drawColor}{RGB}{190,190,190}

\path[draw=drawColor,line width= 0.6pt,line join=round] ( 81.49, 13.51) -- ( 81.49, 62.95);

\path[draw=drawColor,line width= 0.6pt,line join=round] (100.86, 13.51) -- (100.86, 62.95);
\definecolor{drawColor}{RGB}{0,0,0}

\node[text=drawColor,anchor=base,inner sep=0pt, outer sep=0pt, scale=  1.10] at ( 91.17, 54.62) {$\phi$};
\definecolor{drawColor}{RGB}{190,190,190}

\path[draw=drawColor,line width= 0.6pt,line join=round] (125.06, 13.51) -- (125.06, 62.95);
\definecolor{drawColor}{RGB}{0,0,0}

\node[text=drawColor,anchor=base,inner sep=0pt, outer sep=0pt, scale=  1.10] at (112.96, 54.62) {$D$};
\end{scope}
\begin{scope}
\path[clip] (  0.00,  0.00) rectangle (240.90, 67.45);
\definecolor{drawColor}{gray}{0.30}

\node[text=drawColor,anchor=base east,inner sep=0pt, outer sep=0pt, scale=  0.72] at ( 19.35, 24.51) {0};
\end{scope}
\begin{scope}
\path[clip] (  0.00,  0.00) rectangle (240.90, 67.45);
\definecolor{drawColor}{gray}{0.20}

\path[draw=drawColor,line width= 0.6pt,line join=round] ( 21.15, 26.99) --
	( 23.40, 26.99);
\end{scope}
\begin{scope}
\path[clip] (  0.00,  0.00) rectangle (240.90, 67.45);
\definecolor{drawColor}{gray}{0.20}

\path[draw=drawColor,line width= 0.6pt,line join=round] ( 33.08, 11.26) --
	( 33.08, 13.51);

\path[draw=drawColor,line width= 0.6pt,line join=round] ( 81.49, 11.26) --
	( 81.49, 13.51);

\path[draw=drawColor,line width= 0.6pt,line join=round] (129.90, 11.26) --
	(129.90, 13.51);

\path[draw=drawColor,line width= 0.6pt,line join=round] (178.31, 11.26) --
	(178.31, 13.51);

\path[draw=drawColor,line width= 0.6pt,line join=round] (226.72, 11.26) --
	(226.72, 13.51);
\end{scope}
\begin{scope}
\path[clip] (  0.00,  0.00) rectangle (240.90, 67.45);
\definecolor{drawColor}{gray}{0.30}

\node[text=drawColor,anchor=base,inner sep=0pt, outer sep=0pt, scale=  0.72] at ( 33.08,  4.50) {-0.5};

\node[text=drawColor,anchor=base,inner sep=0pt, outer sep=0pt, scale=  0.72] at ( 81.49,  4.50) {0.0};

\node[text=drawColor,anchor=base,inner sep=0pt, outer sep=0pt, scale=  0.72] at (129.90,  4.50) {0.5};

\node[text=drawColor,anchor=base,inner sep=0pt, outer sep=0pt, scale=  0.72] at (178.31,  4.50) {1.0};

\node[text=drawColor,anchor=base,inner sep=0pt, outer sep=0pt, scale=  0.72] at (226.72,  4.50) {1.5};
\end{scope}
\begin{scope}
\path[clip] (  0.00,  0.00) rectangle (240.90, 67.45);
\definecolor{drawColor}{RGB}{0,0,0}

\node[text=drawColor,rotate= 90.00,anchor=base,inner sep=0pt, outer sep=0pt, scale=  0.90] at ( 10.70, 38.23) {$SP$};
\end{scope}
\end{tikzpicture}
\caption{Joint set point ($SP$) as a function of $t$}
\label{fig:control}
\begin{tikzpicture}[x=1pt,y=1pt]
\definecolor{fillColor}{RGB}{255,255,255}
\path[use as bounding box,fill=fillColor,fill opacity=0.00] (0,0) rectangle (240.90, 53.00);
\begin{scope}
\path[clip] ( 23.40, 13.51) rectangle (236.40, 48.50);
\definecolor{drawColor}{RGB}{0,0,0}

\path[draw=drawColor,line width= 0.6pt,line join=round] ( 33.08, 31.00) --
	( 57.29, 15.10) --
	(105.70, 46.91) --
	(154.11, 15.10) --
	(202.51, 46.91) --
	(226.72, 31.00);
\end{scope}
\begin{scope}
\path[clip] (  0.00,  0.00) rectangle (240.90, 53.00);
\definecolor{drawColor}{gray}{0.30}

\node[text=drawColor,anchor=base east,inner sep=0pt, outer sep=0pt, scale=  0.72] at ( 19.35, 12.62) {0};

\node[text=drawColor,anchor=base east,inner sep=0pt, outer sep=0pt, scale=  0.72] at ( 19.35, 44.43) {1};
\end{scope}
\begin{scope}
\path[clip] (  0.00,  0.00) rectangle (240.90, 53.00);
\definecolor{drawColor}{gray}{0.20}

\path[draw=drawColor,line width= 0.6pt,line join=round] ( 21.15, 15.10) --
	( 23.40, 15.10);

\path[draw=drawColor,line width= 0.6pt,line join=round] ( 21.15, 46.91) --
	( 23.40, 46.91);
\end{scope}
\begin{scope}
\path[clip] (  0.00,  0.00) rectangle (240.90, 53.00);
\definecolor{drawColor}{gray}{0.20}

\path[draw=drawColor,line width= 0.6pt,line join=round] ( 57.29, 11.26) --
	( 57.29, 13.51);

\path[draw=drawColor,line width= 0.6pt,line join=round] (105.70, 11.26) --
	(105.70, 13.51);

\path[draw=drawColor,line width= 0.6pt,line join=round] (154.11, 11.26) --
	(154.11, 13.51);

\path[draw=drawColor,line width= 0.6pt,line join=round] (202.51, 11.26) --
	(202.51, 13.51);
\end{scope}
\begin{scope}
\path[clip] (  0.00,  0.00) rectangle (240.90, 53.00);
\definecolor{drawColor}{gray}{0.30}

\node[text=drawColor,anchor=base,inner sep=0pt, outer sep=0pt, scale=  0.72] at ( 57.29,  4.50) {0};

\node[text=drawColor,anchor=base,inner sep=0pt, outer sep=0pt, scale=  0.72] at (105.70,  4.50) {1};

\node[text=drawColor,anchor=base,inner sep=0pt, outer sep=0pt, scale=  0.72] at (154.11,  4.50) {2};

\node[text=drawColor,anchor=base,inner sep=0pt, outer sep=0pt, scale=  0.72] at (202.51,  4.50) {3};
\end{scope}
\begin{scope}
\path[clip] (  0.00,  0.00) rectangle (240.90, 53.00);
\definecolor{drawColor}{RGB}{0,0,0}

\node[text=drawColor,rotate= 90.00,anchor=base,inner sep=0pt, outer sep=0pt, scale=  0.90] at ( 10.70, 31.00) {$D$};
\end{scope}
\end{tikzpicture}
\caption{Wrapping $d$ onto $\left[0, 1\right]$}
\label{fig:dwrap}
\end{figure}

\subsection{Behavior and Fitness}\label{method:features}
The behavior descriptor is composed of two features that are designed to work well as control parameters for a higher-level control system: average turn rate and adjusted forward speed. Average turn speed is measured simply as $b_T = \sum_{i=0}^N \Delta\psi_i/{\Delta t}$, where $N$ is the number of sample periods, $\Delta\psi_i$ is the change in orientation at sample $i$ and $\Delta t$ is the time elapsed. 

\begin{figure}
\begin{tikzpicture}[x=1pt,y=1pt]
\definecolor{fillColor}{RGB}{255,255,255}
\path[use as bounding box,fill=fillColor,fill opacity=0.00] (0,0) rectangle (240.90, 80.30);
\begin{scope}
\path[clip] (  6.75,  6.75) rectangle (236.40, 75.80);
\definecolor{fillColor}{RGB}{0,192,255}

\path[fill=fillColor,fill opacity=0.10] (225.82, 14.61) --
	(214.87, 24.44) --
	(204.42, 39.20) --
	(195.63, 32.28) --
	(183.74, 40.16) --
	(173.54, 41.01) --
	(158.70, 60.09) --
	(148.80, 50.19) --
	(138.13, 52.86) --
	(123.26, 53.81) --
	(104.69, 66.12) --
	( 95.59, 54.70) --
	( 82.38, 57.60) --
	( 70.13, 55.35) --
	( 46.74, 63.59) --
	( 39.57, 51.26) --
	( 24.88, 47.78) --
	( 57.86, 10.06) --
	( 66.76, 17.08) --
	( 75.94, 23.73) --
	( 85.39, 30.00) --
	( 95.10, 35.88) --
	(105.05, 41.35) --
	(115.22, 46.40) --
	(125.59, 51.04) --
	(136.16, 55.24) --
	(146.89, 59.01) --
	(157.77, 62.33) --
	(168.78, 65.20) --
	(179.90, 67.62) --
	(191.12, 69.58) --
	(202.40, 71.07) --
	(213.75, 72.10) --
	(225.12, 72.66) --
	cycle;
\definecolor{drawColor}{RGB}{0,192,255}

\path[draw=drawColor,line width= 1.1pt,line join=round] (225.12, 72.66) --
	(213.75, 72.10) --
	(202.40, 71.07) --
	(191.12, 69.58) --
	(179.90, 67.62) --
	(168.78, 65.20) --
	(157.77, 62.33) --
	(146.89, 59.01) --
	(136.16, 55.24) --
	(125.59, 51.04) --
	(115.22, 46.40) --
	(105.05, 41.35) --
	( 95.10, 35.88) --
	( 85.39, 30.00) --
	( 75.94, 23.73) --
	( 66.76, 17.08) --
	( 57.86, 10.06);
\definecolor{drawColor}{RGB}{0,0,0}

\path[draw=drawColor,line width= 1.1pt,line join=round] (225.96, 19.33) -- (225.68,  9.89);

\path[draw=drawColor,line width= 1.1pt,line join=round] (183.09, 44.84) -- (184.40, 35.48);

\path[draw=drawColor,line width= 1.1pt,line join=round] (136.70, 57.37) -- (139.56, 48.36);

\path[draw=drawColor,line width= 1.1pt,line join=round] ( 80.22, 61.81) -- ( 84.55, 53.39);

\path[draw=drawColor,line width= 1.1pt,line join=round] ( 22.04, 51.58) -- ( 27.73, 43.99);

\path[draw=drawColor,line width= 0.6pt,line join=round] (225.82, 14.61) -- (216.25, 14.89);

\path[draw=drawColor,line width= 0.6pt,line join=round] (219.78, 16.79) --
	(216.25, 14.89) --
	(219.66, 12.79);

\path[draw=drawColor,line width= 0.6pt,line join=round] (214.87, 24.44) -- (205.37, 25.58);

\path[draw=drawColor,line width= 0.6pt,line join=round] (209.05, 27.15) --
	(205.37, 25.58) --
	(208.57, 23.18);

\path[draw=drawColor,line width= 0.6pt,line join=round] (204.42, 39.20) -- (194.99, 37.59);

\path[draw=drawColor,line width= 0.6pt,line join=round] (198.07, 40.14) --
	(194.99, 37.59) --
	(198.74, 36.20);

\path[draw=drawColor,line width= 0.6pt,line join=round] (195.63, 32.28) -- (186.63, 29.09);

\path[draw=drawColor,line width= 0.6pt,line join=round] (189.22, 32.13) --
	(186.63, 29.09) --
	(190.56, 28.36);

\path[draw=drawColor,line width= 0.6pt,line join=round] (183.74, 40.16) -- (174.27, 38.87);

\path[draw=drawColor,line width= 0.6pt,line join=round] (177.43, 41.32) --
	(174.27, 38.87) --
	(177.97, 37.36);

\path[draw=drawColor,line width= 0.6pt,line join=round] (173.54, 41.01) -- (163.98, 40.57);

\path[draw=drawColor,line width= 0.6pt,line join=round] (167.35, 42.73) --
	(163.98, 40.57) --
	(167.53, 38.73);

\path[draw=drawColor,line width= 0.6pt,line join=round] (158.70, 60.09) -- (149.67, 56.96);

\path[draw=drawColor,line width= 0.6pt,line join=round] (152.29, 59.98) --
	(149.67, 56.96) --
	(153.60, 56.20);

\path[draw=drawColor,line width= 0.6pt,line join=round] (148.80, 50.19) -- (140.46, 45.56);

\path[draw=drawColor,line width= 0.6pt,line join=round] (142.52, 48.99) --
	(140.46, 45.56) --
	(144.46, 45.49);

\path[draw=drawColor,line width= 0.6pt,line join=round] (138.13, 52.86) -- (129.00, 50.04);

\path[draw=drawColor,line width= 0.6pt,line join=round] (131.72, 52.97) --
	(129.00, 50.04) --
	(132.90, 49.15);

\path[draw=drawColor,line width= 0.6pt,line join=round] (123.26, 53.81) -- (113.91, 51.82);

\path[draw=drawColor,line width= 0.6pt,line join=round] (116.88, 54.50) --
	(113.91, 51.82) --
	(117.71, 50.59);

\path[draw=drawColor,line width= 0.6pt,line join=round] (104.69, 66.12) -- ( 96.31, 61.55);

\path[draw=drawColor,line width= 0.6pt,line join=round] ( 98.40, 64.97) --
	( 96.31, 61.55) --
	(100.31, 61.45);

\path[draw=drawColor,line width= 0.6pt,line join=round] ( 95.59, 54.70) -- ( 88.14, 48.77);

\path[draw=drawColor,line width= 0.6pt,line join=round] ( 89.61, 52.50) --
	( 88.14, 48.77) --
	( 92.10, 49.37);

\path[draw=drawColor,line width= 0.6pt,line join=round] ( 82.38, 57.60) -- ( 73.85, 53.32);

\path[draw=drawColor,line width= 0.6pt,line join=round] ( 76.06, 56.66) --
	( 73.85, 53.32) --
	( 77.85, 53.09);

\path[draw=drawColor,line width= 0.6pt,line join=round] ( 70.13, 55.35) -- ( 61.24, 51.85);

\path[draw=drawColor,line width= 0.6pt,line join=round] ( 63.73, 54.98) --
	( 61.24, 51.85) --
	( 65.20, 51.26);

\path[draw=drawColor,line width= 0.6pt,line join=round] ( 46.74, 63.59) -- ( 39.25, 57.71);

\path[draw=drawColor,line width= 0.6pt,line join=round] ( 40.75, 61.42) --
	( 39.25, 57.71) --
	( 43.21, 58.28);

\path[draw=drawColor,line width= 0.6pt,line join=round] ( 39.57, 51.26) -- ( 33.23, 44.20);

\path[draw=drawColor,line width= 0.6pt,line join=round] ( 34.05, 48.11) --
	( 33.23, 44.20) --
	( 37.03, 45.44);

\path[draw=drawColor,line width= 0.6pt,line join=round] ( 24.88, 47.78) -- ( 17.19, 42.17);

\path[draw=drawColor,line width= 0.6pt,line join=round] ( 18.81, 45.82) --
	( 17.19, 42.17) --
	( 21.17, 42.59);
\definecolor{fillColor}{RGB}{0,0,0}

\path[draw=drawColor,line width= 0.4pt,line join=round,line cap=round,fill=fillColor] (225.82, 14.61) circle (  1.43);

\path[draw=drawColor,line width= 0.4pt,line join=round,line cap=round,fill=fillColor] (214.87, 24.44) circle (  1.43);

\path[draw=drawColor,line width= 0.4pt,line join=round,line cap=round,fill=fillColor] (204.42, 39.20) circle (  1.43);

\path[draw=drawColor,line width= 0.4pt,line join=round,line cap=round,fill=fillColor] (195.63, 32.28) circle (  1.43);

\path[draw=drawColor,line width= 0.4pt,line join=round,line cap=round,fill=fillColor] (183.74, 40.16) circle (  1.43);

\path[draw=drawColor,line width= 0.4pt,line join=round,line cap=round,fill=fillColor] (173.54, 41.01) circle (  1.43);

\path[draw=drawColor,line width= 0.4pt,line join=round,line cap=round,fill=fillColor] (158.70, 60.09) circle (  1.43);

\path[draw=drawColor,line width= 0.4pt,line join=round,line cap=round,fill=fillColor] (148.80, 50.19) circle (  1.43);

\path[draw=drawColor,line width= 0.4pt,line join=round,line cap=round,fill=fillColor] (138.13, 52.86) circle (  1.43);

\path[draw=drawColor,line width= 0.4pt,line join=round,line cap=round,fill=fillColor] (123.26, 53.81) circle (  1.43);

\path[draw=drawColor,line width= 0.4pt,line join=round,line cap=round,fill=fillColor] (104.69, 66.12) circle (  1.43);

\path[draw=drawColor,line width= 0.4pt,line join=round,line cap=round,fill=fillColor] ( 95.59, 54.70) circle (  1.43);

\path[draw=drawColor,line width= 0.4pt,line join=round,line cap=round,fill=fillColor] ( 82.38, 57.60) circle (  1.43);

\path[draw=drawColor,line width= 0.4pt,line join=round,line cap=round,fill=fillColor] ( 70.13, 55.35) circle (  1.43);

\path[draw=drawColor,line width= 0.4pt,line join=round,line cap=round,fill=fillColor] ( 46.74, 63.59) circle (  1.43);

\path[draw=drawColor,line width= 0.4pt,line join=round,line cap=round,fill=fillColor] ( 39.57, 51.26) circle (  1.43);

\path[draw=drawColor,line width= 0.4pt,line join=round,line cap=round,fill=fillColor] ( 24.88, 47.78) circle (  1.43);

\path[draw=drawColor,line width= 0.6pt,line join=round] (201.35, 80.30) -- (210.54,  0.00);

\node[text=drawColor,anchor=base west,inner sep=0pt, outer sep=0pt, scale=  1.10] at (205.14, 57.67) {$n_a$};

\path[draw=drawColor,line width= 0.6pt,line join=round] ( 35.65, 80.30) -- ( 92.84,  0.00);

\node[text=drawColor,anchor=base west,inner sep=0pt, outer sep=0pt, scale=  1.10] at ( 87.11, 10.42) {$n_b$};
\end{scope}
\end{tikzpicture}
\vspace{-18pt}
\caption{Forward movement estimation. 
The length of the circular segment is $R\Theta$, which gives an estimate of the amount of forward movement. $SE_R$ approximately corresponds to the shaded area, and measures the amount of sideways motion. }
\label{fig:movement}
\end{figure}
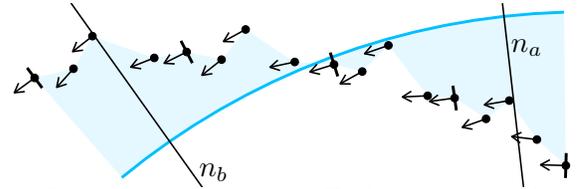

To provide a robust measure of the forward speed that accounts for turn radius and filters out sideways locomotion, more complex calculations are carried out for the second behavior feature: First the average position and orientation of the robot during the first and last periods are estimated. Lines perpendicular to the orientations and intersecting the positions are constructed, illustrated as $n_a$ and $n_b$ in \autoref{fig:movement}. The point where these two lines cross is used as a center of curvature, $C$; using it we can estimate the turn radius and its standard error
\begin{align*}
  R &= \textstyle\frac{1}{N+1}\sum_{i=0}^N \left|x_i-C\right| \\
 SE_R &= \textstyle\sqrt{\frac{1}{N}\sum_{i=0}^N \paren{\left|x_i-C\right| - R}^2}
\end{align*}
and the change in orientation relative to $C$
\[ \textstyle \Theta = \frac{1}{N}\sum_{i=1}^N \angle x_iCx_{i-1} \]
where then angle $\angle x_iCx_{i-1}$ between $x_iC$ and $Cx_{i-1}$ signed. The adjusted forward speed is then defined as $b_F = R\Theta/{\Delta t}$.

As the fitness measure, we compute a weighted sum of several penalty measures
\[ \op{Penalty} = 10SE_R + \overline{|\theta|} + 10^{-\overline{h}/0.05} \]
where $\overline{|\theta|}$ is the average absolute pitch of the body, and $\overline{h}$ is the average height of the body reference point. This penalizes large sideways motion, up/down tilt and having the body close to the ground. The penalty score is then inverted into
\[ f = \max\left\lbrace 0, -5\log_{10}\op{Penalty} \right\rbrace \]
in order to produce a value increasing with solution quality.

\subsection{Mutation Operators and Other Variables}\label{method:mutators}
We test two different commonly used mutation operators: mutating all parameters, and mutating each parameter with a probability $1/k$. In both cases Gaussian mutation with standard deviation $\sigma$ is used on the mutated parameters. We run the experiments with five different values for $\sigma$. We also attempt to control for effects causesd by running with three different map resolutions. The extents the map cells covers in behavior space are kept constant.
All combinations of mutation type, mutation magnitude and map resolution is run multiple times on all robots, as well as map extents is summarized in \autoref{tbl:setup}.

\subsection{Performance Measures}\label{method:meas}
\label{sec:perfmeas}

Performance measures similar to those in \cite{Mouret2015} and \cite{Pugh2016} are used:
\Precision is defined as the average score of filled cells: 
\[ \opp Pm = \frac{\opp{QD-score}{m}}{\opp nm} = \frac{1}{\opp nm}\sum_{\vx \in M}m\paren{\vx} \]
where $m\paren{x}$ is the score in cell $x$ of map $m$ or zero if empty, 
$M$ is the set of all cells,
and $\opp{n}{m}$ is the number of filled cells in $m$.
\Coverage is defined as the fraction of cells filled:
\[ \opp Cm = \frac{\opp nm}{\opp Nm} \]
where $\opp Nm$ is the total number of cells in the map. 
Finally, global reliability is defined as the average score across all cells, which can be expressed as
\[ \opp Gm = \frac{\opp{QD-score}{m}}{\opp Nm} = \opp Pm\opp Cm \]
meaning that reliability can be seen as either the QD-score normalized for map resolution, or the product of \precision and \coverage. These definitions differs from those in \cite{Mouret2015} in that they do not scale the cell scores against the best observed score for each cell.

\subsection{Cliff's Delta}\label{method:effect-size}
Cliff's delta \cite{Cliff1993} is an ordinal effect size that estimates how separated two distributions are. Given the probability \hbox{$P_A = P(a>b)$} of a random value from group $a$ being greater than a random value of group $b$, and $P_B = P(a<b)$, it can be defined as
\begin{equation} \label{eq:d}
\delta_{ab} = P(a > b) - P(a < b)
\end{equation}

\begin{table}
\caption{Example values for $\delta_{ab}$}
\centering\small
\begin{tabular}{ccl@{\hskip 3pt}ll@{\hskip 6pt}l@{\hskip 6pt}l}
              &  &  \multicolumn{2}{c}{No draws} & \multicolumn{3}{c}{Max. draws}\\
$\delta_{ab}$ & $P_A / P_B$ & $P_A$ & $P_B$ & $P_A$ & $P_B$ & $P_D$ \\
\hline
	$+1.0$           & $\infty$           & 1.0  & 0.0  & 1.0 & 0.0 & 0.0 \\
	$+0.5$           & At least $3:1$     & 0.75 & 0.25 & 0.5 & 0.0 & 0.5 \\
	$\phantom{+}0.0$ & Undefined or $1:1$ & 0.5  & 0.5  & 0.0 & 0.0 & 1.0 \\
	$-0.2$           & At most $2:3$      & 0.4  & 0.6  & 0.0 & 0.2 & 0.8 \\
	$-0.8$           & At most $1:9$      & 0.1  & 0.9  & 0.0 & 0.8 & 0.2 \\
	$-1.0$           & $0$                & 0.0  & 1.0  & 0.0 & 1.0 & 0.0 \\
\hline
\end{tabular}
\label{tbl:deltaex}
\end{table}

\autoref{tbl:deltaex} shows some example values of Cliff's delta, along with an interpretation in terms of the probabilities \hbox{$P_A = P(a > b)$} and $P_B = P(a < b)$ along with the probability of a draw $P_D = 1-(P_A+P_B)$. The sign of $\delta_{ab}$ says whether $a$ (positive) or $b$ (negative) is more likely to perform the best, and the magnitude indicates the degree of difference in probabilities.

Cliff's delta can either be calculated exactly by comparing all possible pairs of observations from the samples, approximated by comparing a large number of random pairs, or through its relation 
\[ \delta_{ab} = \frac{2U}{|a||b|} - 1 \]
to the Wilcoxon-Mann-Whitney (WMW) U statistic. Because of this linear relation, testing whether $\delta_{ab}$ is likely to be zero corresponds to doing a WMW U test. $\delta_{ab}$ is also related to Vargha and Delaney's effect size $\hat{A}$ \cite{Vargha2000} by $\delta_{ab} = 2\hat{A}_{ab} - 1$.

Since it is ordinal, we can use it to measure differences across incomparable groups, for example compare the performance of two algorithms across different benchmarks, or in this case, different parametrizations of an algorithm across different robot morphologies. Assuming the different groups are of equal importance, this can be done by ensuring that comparisons are always done between observations in the same group, and weighting the groups equally. This can be calculated as
\[ \delta_{ab} = \frac{1}{|G|}\sum_{g\in G} \delta\paren{a\,|\,g,\ b\,|\,g} \]

\begin{figure*}

\setlength\tabcolsep{0pt}
\begin{tabular}{ccc}
\begin{tikzpicture}[x=1pt,y=1pt]
\definecolor{fillColor}{RGB}{255,255,255}
\path[use as bounding box,fill=fillColor,fill opacity=0.00] (0,0) rectangle (166.22,144.54);
\begin{scope}
\path[clip] ( 17.75, 24.19) rectangle (147.76,140.04);
\definecolor{drawColor}{gray}{0.20}

\path[draw=drawColor,line width= 0.6pt,line join=round] ( 25.40, 46.87) -- ( 25.40, 68.25);

\path[draw=drawColor,line width= 0.6pt,line join=round] ( 25.40, 39.31) -- ( 25.40, 32.39);
\definecolor{fillColor}{RGB}{0,192,0}

\path[draw=drawColor,line width= 0.6pt,line join=round,line cap=round,fill=fillColor,fill opacity=0.60] ( 21.57, 46.87) --
	( 21.57, 39.31) --
	( 29.22, 39.31) --
	( 29.22, 46.87) --
	( 21.57, 46.87) --
	cycle;

\path[draw=drawColor,line width= 1.1pt,line join=round] ( 21.57, 43.22) -- ( 29.22, 43.22);

\path[draw=drawColor,line width= 0.6pt,line join=round] ( 38.14, 57.17) -- ( 38.14, 73.80);

\path[draw=drawColor,line width= 0.6pt,line join=round] ( 38.14, 40.88) -- ( 38.14, 29.45);
\definecolor{fillColor}{RGB}{0,192,255}

\path[draw=drawColor,line width= 0.6pt,line join=round,line cap=round,fill=fillColor,fill opacity=0.60] ( 34.32, 57.17) --
	( 34.32, 40.88) --
	( 41.97, 40.88) --
	( 41.97, 57.17) --
	( 34.32, 57.17) --
	cycle;

\path[draw=drawColor,line width= 1.1pt,line join=round] ( 34.32, 47.20) -- ( 41.97, 47.20);

\path[draw=drawColor,line width= 0.6pt,line join=round] ( 50.89, 66.42) -- ( 50.89, 90.63);

\path[draw=drawColor,line width= 0.6pt,line join=round] ( 50.89, 49.36) -- ( 50.89, 38.05);
\definecolor{fillColor}{RGB}{160,96,255}

\path[draw=drawColor,line width= 0.6pt,line join=round,line cap=round,fill=fillColor,fill opacity=0.60] ( 47.06, 66.42) --
	( 47.06, 49.36) --
	( 54.71, 49.36) --
	( 54.71, 66.42) --
	( 47.06, 66.42) --
	cycle;

\path[draw=drawColor,line width= 1.1pt,line join=round] ( 47.06, 54.86) -- ( 54.71, 54.86);

\path[draw=drawColor,line width= 0.6pt,line join=round] ( 63.64, 84.88) -- ( 63.64,109.45);

\path[draw=drawColor,line width= 0.6pt,line join=round] ( 63.64, 64.97) -- ( 63.64, 40.54);
\definecolor{fillColor}{RGB}{255,32,128}

\path[draw=drawColor,line width= 0.6pt,line join=round,line cap=round,fill=fillColor,fill opacity=0.60] ( 59.81, 84.88) --
	( 59.81, 64.97) --
	( 67.46, 64.97) --
	( 67.46, 84.88) --
	( 59.81, 84.88) --
	cycle;

\path[draw=drawColor,line width= 1.1pt,line join=round] ( 59.81, 73.80) -- ( 67.46, 73.80);

\path[draw=drawColor,line width= 0.6pt,line join=round] ( 76.38, 97.48) -- ( 76.38,134.77);

\path[draw=drawColor,line width= 0.6pt,line join=round] ( 76.38, 74.56) -- ( 76.38, 49.36);
\definecolor{fillColor}{RGB}{255,176,0}

\path[draw=drawColor,line width= 0.6pt,line join=round,line cap=round,fill=fillColor,fill opacity=0.60] ( 72.56, 97.48) --
	( 72.56, 74.56) --
	( 80.21, 74.56) --
	( 80.21, 97.48) --
	( 72.56, 97.48) --
	cycle;

\path[draw=drawColor,line width= 1.1pt,line join=round] ( 72.56, 86.67) -- ( 80.21, 86.67);

\path[draw=drawColor,line width= 0.6pt,line join=round] ( 89.13, 73.80) -- ( 89.13,134.77);

\path[draw=drawColor,line width= 0.6pt,line join=round] ( 89.13, 51.63) -- ( 89.13, 29.45);
\definecolor{fillColor}{RGB}{0,192,0}

\path[draw=drawColor,line width= 0.6pt,line join=round,line cap=round,fill=fillColor,fill opacity=0.60] ( 85.30, 73.80) --
	( 85.30, 51.63) --
	( 92.95, 51.63) --
	( 92.95, 73.80) --
	( 85.30, 73.80) --
	cycle;

\path[draw=drawColor,line width= 1.1pt,line join=round] ( 85.30, 68.25) -- ( 92.95, 68.25);

\path[draw=drawColor,line width= 0.6pt,line join=round] (101.87, 74.82) -- (101.87,111.58);

\path[draw=drawColor,line width= 0.6pt,line join=round] (101.87, 46.53) -- (101.87, 32.39);
\definecolor{fillColor}{RGB}{0,192,255}

\path[draw=drawColor,line width= 0.6pt,line join=round,line cap=round,fill=fillColor,fill opacity=0.60] ( 98.05, 74.82) --
	( 98.05, 46.53) --
	(105.70, 46.53) --
	(105.70, 74.82) --
	( 98.05, 74.82) --
	cycle;

\path[draw=drawColor,line width= 1.1pt,line join=round] ( 98.05, 55.02) -- (105.70, 55.02);

\path[draw=drawColor,line width= 0.6pt,line join=round] (114.62, 80.80) -- (114.62,109.45);

\path[draw=drawColor,line width= 0.6pt,line join=round] (114.62, 42.30) -- (114.62, 32.46);
\definecolor{fillColor}{RGB}{160,96,255}

\path[draw=drawColor,line width= 0.6pt,line join=round,line cap=round,fill=fillColor,fill opacity=0.60] (110.80, 80.80) --
	(110.80, 42.30) --
	(118.45, 42.30) --
	(118.45, 80.80) --
	(110.80, 80.80) --
	cycle;

\path[draw=drawColor,line width= 1.1pt,line join=round] (110.80, 54.70) -- (118.45, 54.70);

\path[draw=drawColor,line width= 0.6pt,line join=round] (127.37, 82.08) -- (127.37,134.77);

\path[draw=drawColor,line width= 0.6pt,line join=round] (127.37, 46.13) -- (127.37, 29.45);
\definecolor{fillColor}{RGB}{0,192,0}

\path[draw=drawColor,line width= 0.6pt,line join=round,line cap=round,fill=fillColor,fill opacity=0.60] (123.54, 82.08) --
	(123.54, 46.13) --
	(131.19, 46.13) --
	(131.19, 82.08) --
	(123.54, 82.08) --
	cycle;

\path[draw=drawColor,line width= 1.1pt,line join=round] (123.54, 57.17) -- (131.19, 57.17);

\path[draw=drawColor,line width= 0.6pt,line join=round] (140.11, 73.80) -- (140.11,111.58);

\path[draw=drawColor,line width= 0.6pt,line join=round] (140.11, 46.53) -- (140.11, 32.39);
\definecolor{fillColor}{RGB}{0,192,255}

\path[draw=drawColor,line width= 0.6pt,line join=round,line cap=round,fill=fillColor,fill opacity=0.60] (136.29, 73.80) --
	(136.29, 46.53) --
	(143.94, 46.53) --
	(143.94, 73.80) --
	(136.29, 73.80) --
	cycle;

\path[draw=drawColor,line width= 1.1pt,line join=round] (136.29, 60.68) -- (143.94, 60.68);
\definecolor{drawColor}{RGB}{0,0,0}

\path[draw=drawColor,draw opacity=0.50,line width= 0.6pt,line join=round] ( 82.75, 24.19) -- ( 82.75,140.04);

\path[draw=drawColor,draw opacity=0.50,line width= 0.6pt,line join=round] (120.99, 24.19) -- (120.99,140.04);
\end{scope}
\begin{scope}
\path[clip] (147.76, 24.19) rectangle (161.72,140.04);
\definecolor{drawColor}{gray}{0.10}

\node[text=drawColor,rotate=-90.00,anchor=base,inner sep=0pt, outer sep=0pt, scale=  0.72] at (152.26, 82.11) {robot3};
\end{scope}
\begin{scope}
\path[clip] (  0.00,  0.00) rectangle (166.22,144.54);
\definecolor{drawColor}{gray}{0.20}

\path[draw=drawColor,line width= 0.6pt,line join=round] ( 25.40, 21.94) --
	( 25.40, 24.19);

\path[draw=drawColor,line width= 0.6pt,line join=round] ( 38.14, 21.94) --
	( 38.14, 24.19);

\path[draw=drawColor,line width= 0.6pt,line join=round] ( 50.89, 21.94) --
	( 50.89, 24.19);

\path[draw=drawColor,line width= 0.6pt,line join=round] ( 63.64, 21.94) --
	( 63.64, 24.19);

\path[draw=drawColor,line width= 0.6pt,line join=round] ( 76.38, 21.94) --
	( 76.38, 24.19);

\path[draw=drawColor,line width= 0.6pt,line join=round] ( 89.13, 21.94) --
	( 89.13, 24.19);

\path[draw=drawColor,line width= 0.6pt,line join=round] (101.87, 21.94) --
	(101.87, 24.19);

\path[draw=drawColor,line width= 0.6pt,line join=round] (114.62, 21.94) --
	(114.62, 24.19);

\path[draw=drawColor,line width= 0.6pt,line join=round] (127.37, 21.94) --
	(127.37, 24.19);

\path[draw=drawColor,line width= 0.6pt,line join=round] (140.11, 21.94) --
	(140.11, 24.19);
\end{scope}
\begin{scope}
\path[clip] (  0.00,  0.00) rectangle (166.22,144.54);
\definecolor{drawColor}{gray}{0.30}

\node[text=drawColor,rotate= 90.00,anchor=base,inner sep=0pt, outer sep=0pt, scale=  0.72] at ( 30.35, 12.32) {0.05};

\node[text=drawColor,rotate= 90.00,anchor=base,inner sep=0pt, outer sep=0pt, scale=  0.72] at ( 43.10, 12.32) {0.1};

\node[text=drawColor,rotate= 90.00,anchor=base,inner sep=0pt, outer sep=0pt, scale=  0.72] at ( 55.85, 12.32) {0.2};

\node[text=drawColor,rotate= 90.00,anchor=base,inner sep=0pt, outer sep=0pt, scale=  0.72] at ( 68.59, 12.32) {0.4};

\node[text=drawColor,rotate= 90.00,anchor=base,inner sep=0pt, outer sep=0pt, scale=  0.72] at ( 81.34, 12.32) {0.8};

\node[text=drawColor,rotate= 90.00,anchor=base,inner sep=0pt, outer sep=0pt, scale=  0.72] at ( 94.09, 12.32) {5$\times$5};

\node[text=drawColor,rotate= 90.00,anchor=base,inner sep=0pt, outer sep=0pt, scale=  0.72] at (106.83, 12.32) {7$\times$7};

\node[text=drawColor,rotate= 90.00,anchor=base,inner sep=0pt, outer sep=0pt, scale=  0.72] at (119.58, 12.32) {9$\times$9};

\node[text=drawColor,rotate= 90.00,anchor=base,inner sep=0pt, outer sep=0pt, scale=  0.72] at (132.33, 12.32) {all};

\node[text=drawColor,rotate= 90.00,anchor=base,inner sep=0pt, outer sep=0pt, scale=  0.72] at (145.07, 12.32) {some};
\end{scope}
\begin{scope}
\path[clip] (  0.00,  0.00) rectangle (166.22,144.54);
\definecolor{drawColor}{gray}{0.30}

\node[text=drawColor, anchor=base,inner sep=0pt, outer sep=0pt, scale=  0.72] at (50.89, 134.77) { Magnitude };
\node[text=drawColor, anchor=base,inner sep=0pt, outer sep=0pt, scale=  0.72] at (101.87, 134.77) { Resolution };
\node[text=drawColor, anchor=base,inner sep=0pt, outer sep=0pt, scale=  0.72] at (133.6, 134.77) { Type };

\node[text=drawColor,anchor=base east,inner sep=0pt, outer sep=0pt, scale=  0.72] at ( 13.70, 49.15) {0.2};

\node[text=drawColor,anchor=base east,inner sep=0pt, outer sep=0pt, scale=  0.72] at ( 13.70, 76.86) {0.4};

\node[text=drawColor,anchor=base east,inner sep=0pt, outer sep=0pt, scale=  0.72] at ( 13.70,104.58) {0.6};

\node[text=drawColor,anchor=base east,inner sep=0pt, outer sep=0pt, scale=  0.72] at ( 13.70,132.29) {0.8};
\end{scope}
\begin{scope}
\path[clip] (  0.00,  0.00) rectangle (166.22,144.54);
\definecolor{drawColor}{gray}{0.20}

\path[draw=drawColor,line width= 0.6pt,line join=round] ( 15.50, 51.63) --
	( 17.75, 51.63);

\path[draw=drawColor,line width= 0.6pt,line join=round] ( 15.50, 79.34) --
	( 17.75, 79.34);

\path[draw=drawColor,line width= 0.6pt,line join=round] ( 15.50,107.06) --
	( 17.75,107.06);

\path[draw=drawColor,line width= 0.6pt,line join=round] ( 15.50,134.77) --
	( 17.75,134.77);

\end{scope}
\end{tikzpicture}
&
\begin{tikzpicture}[x=1pt,y=1pt]
\definecolor{fillColor}{RGB}{255,255,255}
\path[use as bounding box,fill=fillColor,fill opacity=0.00] (0,0) rectangle (166.22,144.54);
\begin{scope}
\path[clip] ( 12.15, 24.19) rectangle (147.76,140.04);
\definecolor{drawColor}{gray}{0.20}

\path[draw=drawColor,line width= 0.6pt,line join=round] ( 20.13, 53.34) -- ( 20.13, 88.19);

\path[draw=drawColor,line width= 0.6pt,line join=round] ( 20.13, 39.89) -- ( 20.13, 29.45);
\definecolor{fillColor}{RGB}{0,192,0}

\path[draw=drawColor,line width= 0.6pt,line join=round,line cap=round,fill=fillColor,fill opacity=0.60] ( 16.14, 53.34) --
	( 16.14, 39.89) --
	( 24.11, 39.89) --
	( 24.11, 53.34) --
	( 16.14, 53.34) --
	cycle;

\path[draw=drawColor,line width= 1.1pt,line join=round] ( 16.14, 46.83) -- ( 24.11, 46.83);

\path[draw=drawColor,line width= 0.6pt,line join=round] ( 33.42, 65.74) -- ( 33.42, 99.12);

\path[draw=drawColor,line width= 0.6pt,line join=round] ( 33.42, 46.07) -- ( 33.42, 32.01);
\definecolor{fillColor}{RGB}{0,192,255}

\path[draw=drawColor,line width= 0.6pt,line join=round,line cap=round,fill=fillColor,fill opacity=0.60] ( 29.43, 65.74) --
	( 29.43, 46.07) --
	( 37.41, 46.07) --
	( 37.41, 65.74) --
	( 29.43, 65.74) --
	cycle;

\path[draw=drawColor,line width= 1.1pt,line join=round] ( 29.43, 55.10) -- ( 37.41, 55.10);

\path[draw=drawColor,line width= 0.6pt,line join=round] ( 46.72, 78.67) -- ( 46.72,107.43);

\path[draw=drawColor,line width= 0.6pt,line join=round] ( 46.72, 56.49) -- ( 46.72, 40.63);
\definecolor{fillColor}{RGB}{160,96,255}

\path[draw=drawColor,line width= 0.6pt,line join=round,line cap=round,fill=fillColor,fill opacity=0.60] ( 42.73, 78.67) --
	( 42.73, 56.49) --
	( 50.71, 56.49) --
	( 50.71, 78.67) --
	( 42.73, 78.67) --
	cycle;

\path[draw=drawColor,line width= 1.1pt,line join=round] ( 42.73, 63.70) -- ( 50.71, 63.70);

\path[draw=drawColor,line width= 0.6pt,line join=round] ( 60.01,101.87) -- ( 60.01,129.37);

\path[draw=drawColor,line width= 0.6pt,line join=round] ( 60.01, 80.69) -- ( 60.01, 41.88);
\definecolor{fillColor}{RGB}{255,32,128}

\path[draw=drawColor,line width= 0.6pt,line join=round,line cap=round,fill=fillColor,fill opacity=0.60] ( 56.02,101.87) --
	( 56.02, 80.69) --
	( 64.00, 80.69) --
	( 64.00,101.87) --
	( 56.02,101.87) --
	cycle;

\path[draw=drawColor,line width= 1.1pt,line join=round] ( 56.02, 93.99) -- ( 64.00, 93.99);

\path[draw=drawColor,line width= 0.6pt,line join=round] ( 73.31,105.74) -- ( 73.31,134.77);

\path[draw=drawColor,line width= 0.6pt,line join=round] ( 73.31, 82.20) -- ( 73.31, 47.87);
\definecolor{fillColor}{RGB}{255,176,0}

\path[draw=drawColor,line width= 0.6pt,line join=round,line cap=round,fill=fillColor,fill opacity=0.60] ( 69.32,105.74) --
	( 69.32, 82.20) --
	( 77.30, 82.20) --
	( 77.30,105.74) --
	( 69.32,105.74) --
	cycle;

\path[draw=drawColor,line width= 1.1pt,line join=round] ( 69.32, 93.41) -- ( 77.30, 93.41);

\path[draw=drawColor,line width= 0.6pt,line join=round] ( 86.60, 99.17) -- ( 86.60,134.77);

\path[draw=drawColor,line width= 0.6pt,line join=round] ( 86.60, 59.40) -- ( 86.60, 32.01);
\definecolor{fillColor}{RGB}{0,192,0}

\path[draw=drawColor,line width= 0.6pt,line join=round,line cap=round,fill=fillColor,fill opacity=0.60] ( 82.61, 99.17) --
	( 82.61, 59.40) --
	( 90.59, 59.40) --
	( 90.59, 99.17) --
	( 82.61, 99.17) --
	cycle;

\path[draw=drawColor,line width= 1.1pt,line join=round] ( 82.61, 82.19) -- ( 90.59, 82.19);

\path[draw=drawColor,line width= 0.6pt,line join=round] ( 99.90, 88.11) -- ( 99.90,129.37);

\path[draw=drawColor,line width= 0.6pt,line join=round] ( 99.90, 53.29) -- ( 99.90, 29.45);
\definecolor{fillColor}{RGB}{0,192,255}

\path[draw=drawColor,line width= 0.6pt,line join=round,line cap=round,fill=fillColor,fill opacity=0.60] ( 95.91, 88.11) --
	( 95.91, 53.29) --
	(103.89, 53.29) --
	(103.89, 88.11) --
	( 95.91, 88.11) --
	cycle;

\path[draw=drawColor,line width= 1.1pt,line join=round] ( 95.91, 61.73) -- (103.89, 61.73);

\path[draw=drawColor,line width= 0.6pt,line join=round] (113.19, 85.37) -- (113.19,116.65);

\path[draw=drawColor,line width= 0.6pt,line join=round] (113.19, 45.16) -- (113.19, 31.45);
\definecolor{fillColor}{RGB}{160,96,255}

\path[draw=drawColor,line width= 0.6pt,line join=round,line cap=round,fill=fillColor,fill opacity=0.60] (109.21, 85.37) --
	(109.21, 45.16) --
	(117.18, 45.16) --
	(117.18, 85.37) --
	(109.21, 85.37) --
	cycle;

\path[draw=drawColor,line width= 1.1pt,line join=round] (109.21, 61.99) -- (117.18, 61.99);

\path[draw=drawColor,line width= 0.6pt,line join=round] (126.49, 90.29) -- (126.49,125.78);

\path[draw=drawColor,line width= 0.6pt,line join=round] (126.49, 51.38) -- (126.49, 29.45);
\definecolor{fillColor}{RGB}{0,192,0}

\path[draw=drawColor,line width= 0.6pt,line join=round,line cap=round,fill=fillColor,fill opacity=0.60] (122.50, 90.29) --
	(122.50, 51.38) --
	(130.48, 51.38) --
	(130.48, 90.29) --
	(122.50, 90.29) --
	cycle;

\path[draw=drawColor,line width= 1.1pt,line join=round] (122.50, 66.22) -- (130.48, 66.22);

\path[draw=drawColor,line width= 0.6pt,line join=round] (139.78, 92.23) -- (139.78,134.77);

\path[draw=drawColor,line width= 0.6pt,line join=round] (139.78, 51.41) -- (139.78, 29.85);
\definecolor{fillColor}{RGB}{0,192,255}

\path[draw=drawColor,line width= 0.6pt,line join=round,line cap=round,fill=fillColor,fill opacity=0.60] (135.80, 92.23) --
	(135.80, 51.41) --
	(143.77, 51.41) --
	(143.77, 92.23) --
	(135.80, 92.23) --
	cycle;

\path[draw=drawColor,line width= 1.1pt,line join=round] (135.80, 70.06) -- (143.77, 70.06);
\definecolor{drawColor}{RGB}{0,0,0}

\path[draw=drawColor,draw opacity=0.50,line width= 0.6pt,line join=round] ( 79.96, 24.19) -- ( 79.96,140.04);

\path[draw=drawColor,draw opacity=0.50,line width= 0.6pt,line join=round] (119.84, 24.19) -- (119.84,140.04);
\end{scope}
\begin{scope}
\path[clip] (147.76, 24.19) rectangle (161.72,140.04);
\definecolor{drawColor}{gray}{0.10}

\node[text=drawColor,rotate=-90.00,anchor=base,inner sep=0pt, outer sep=0pt, scale=  0.72] at (152.26, 82.11) {robot3};
\end{scope}
\begin{scope}
\path[clip] (  0.00,  0.00) rectangle (166.22,144.54);
\definecolor{drawColor}{gray}{0.20}

\path[draw=drawColor,line width= 0.6pt,line join=round] ( 20.13, 21.94) --
	( 20.13, 24.19);

\path[draw=drawColor,line width= 0.6pt,line join=round] ( 33.42, 21.94) --
	( 33.42, 24.19);

\path[draw=drawColor,line width= 0.6pt,line join=round] ( 46.72, 21.94) --
	( 46.72, 24.19);

\path[draw=drawColor,line width= 0.6pt,line join=round] ( 60.01, 21.94) --
	( 60.01, 24.19);

\path[draw=drawColor,line width= 0.6pt,line join=round] ( 73.31, 21.94) --
	( 73.31, 24.19);

\path[draw=drawColor,line width= 0.6pt,line join=round] ( 86.60, 21.94) --
	( 86.60, 24.19);

\path[draw=drawColor,line width= 0.6pt,line join=round] ( 99.90, 21.94) --
	( 99.90, 24.19);

\path[draw=drawColor,line width= 0.6pt,line join=round] (113.19, 21.94) --
	(113.19, 24.19);

\path[draw=drawColor,line width= 0.6pt,line join=round] (126.49, 21.94) --
	(126.49, 24.19);

\path[draw=drawColor,line width= 0.6pt,line join=round] (139.78, 21.94) --
	(139.78, 24.19);
\end{scope}
\begin{scope}
\path[clip] (  0.00,  0.00) rectangle (166.22,144.54);
\definecolor{drawColor}{gray}{0.30}

\node[text=drawColor,rotate= 90.00,anchor=base,inner sep=0pt, outer sep=0pt, scale=  0.72] at ( 25.09, 12.32) {0.05};

\node[text=drawColor,rotate= 90.00,anchor=base,inner sep=0pt, outer sep=0pt, scale=  0.72] at ( 38.38, 12.32) {0.1};

\node[text=drawColor,rotate= 90.00,anchor=base,inner sep=0pt, outer sep=0pt, scale=  0.72] at ( 51.68, 12.32) {0.2};

\node[text=drawColor,rotate= 90.00,anchor=base,inner sep=0pt, outer sep=0pt, scale=  0.72] at ( 64.97, 12.32) {0.4};

\node[text=drawColor,rotate= 90.00,anchor=base,inner sep=0pt, outer sep=0pt, scale=  0.72] at ( 78.27, 12.32) {0.8};

\node[text=drawColor,rotate= 90.00,anchor=base,inner sep=0pt, outer sep=0pt, scale=  0.72] at ( 91.56, 12.32) {5$\times$5};

\node[text=drawColor,rotate= 90.00,anchor=base,inner sep=0pt, outer sep=0pt, scale=  0.72] at (104.86, 12.32) {7$\times$7};

\node[text=drawColor,rotate= 90.00,anchor=base,inner sep=0pt, outer sep=0pt, scale=  0.72] at (118.15, 12.32) {9$\times$9};

\node[text=drawColor,rotate= 90.00,anchor=base,inner sep=0pt, outer sep=0pt, scale=  0.72] at (131.45, 12.32) {all};

\node[text=drawColor,rotate= 90.00,anchor=base,inner sep=0pt, outer sep=0pt, scale=  0.72] at (144.74, 12.32) {some};
\end{scope}
\begin{scope}
\path[clip] (  0.00,  0.00) rectangle (166.22,144.54);
\definecolor{drawColor}{gray}{0.30}

\node[text=drawColor, anchor=base,inner sep=0pt, outer sep=0pt, scale=  0.72] at (49.89, 134.77) { Magnitude };
\node[text=drawColor, anchor=base,inner sep=0pt, outer sep=0pt, scale=  0.72] at (100.87, 134.77) { Resolution };
\node[text=drawColor, anchor=base,inner sep=0pt, outer sep=0pt, scale=  0.72] at (132.6, 134.77) { Type };

\node[text=drawColor,anchor=base east,inner sep=0pt, outer sep=0pt, scale=  0.72] at (  8.10, 56.68) {1};

\node[text=drawColor,anchor=base east,inner sep=0pt, outer sep=0pt, scale=  0.72] at (  8.10,101.64) {2};
\end{scope}
\begin{scope}
\path[clip] (  0.00,  0.00) rectangle (166.22,144.54);
\definecolor{drawColor}{gray}{0.20}

\path[draw=drawColor,line width= 0.6pt,line join=round] (  9.90, 59.16) --
	( 12.15, 59.16);

\path[draw=drawColor,line width= 0.6pt,line join=round] (  9.90,104.12) --
	( 12.15,104.12);
\end{scope}
\end{tikzpicture}
&
\begin{tikzpicture}[x=1pt,y=1pt]
\definecolor{fillColor}{RGB}{255,255,255}
\path[use as bounding box,fill=fillColor,fill opacity=0.00] (0,0) rectangle (166.22,144.54);
\begin{scope}
\path[clip] ( 15.75, 24.19) rectangle (147.76,140.04);
\definecolor{drawColor}{gray}{0.20}

\path[draw=drawColor,line width= 0.6pt,line join=round] ( 23.51, 76.77) -- ( 23.51, 93.70);

\path[draw=drawColor,line width= 0.6pt,line join=round] ( 23.51, 59.09) -- ( 23.51, 43.16);
\definecolor{fillColor}{RGB}{0,192,0}

\path[draw=drawColor,line width= 0.6pt,line join=round,line cap=round,fill=fillColor,fill opacity=0.60] ( 19.63, 76.77) --
	( 19.63, 59.09) --
	( 27.40, 59.09) --
	( 27.40, 76.77) --
	( 19.63, 76.77) --
	cycle;

\path[draw=drawColor,line width= 1.1pt,line join=round] ( 19.63, 66.99) -- ( 27.40, 66.99);

\path[draw=drawColor,line width= 0.6pt,line join=round] ( 36.46, 76.80) -- ( 36.46,134.77);

\path[draw=drawColor,line width= 0.6pt,line join=round] ( 36.46, 59.40) -- ( 36.46, 35.90);
\definecolor{fillColor}{RGB}{0,192,255}

\path[draw=drawColor,line width= 0.6pt,line join=round,line cap=round,fill=fillColor,fill opacity=0.60] ( 32.57, 76.80) --
	( 32.57, 59.40) --
	( 40.34, 59.40) --
	( 40.34, 76.80) --
	( 32.57, 76.80) --
	cycle;

\path[draw=drawColor,line width= 1.1pt,line join=round] ( 32.57, 68.41) -- ( 40.34, 68.41);

\path[draw=drawColor,line width= 0.6pt,line join=round] ( 49.40, 73.15) -- ( 49.40, 93.80);

\path[draw=drawColor,line width= 0.6pt,line join=round] ( 49.40, 60.91) -- ( 49.40, 44.97);
\definecolor{fillColor}{RGB}{160,96,255}

\path[draw=drawColor,line width= 0.6pt,line join=round,line cap=round,fill=fillColor,fill opacity=0.60] ( 45.52, 73.15) --
	( 45.52, 60.91) --
	( 53.28, 60.91) --
	( 53.28, 73.15) --
	( 45.52, 73.15) --
	cycle;

\path[draw=drawColor,line width= 1.1pt,line join=round] ( 45.52, 65.08) -- ( 53.28, 65.08);

\path[draw=drawColor,line width= 0.6pt,line join=round] ( 62.34, 65.77) -- ( 62.34, 81.09);

\path[draw=drawColor,line width= 0.6pt,line join=round] ( 62.34, 53.01) -- ( 62.34, 34.24);
\definecolor{fillColor}{RGB}{255,32,128}

\path[draw=drawColor,line width= 0.6pt,line join=round,line cap=round,fill=fillColor,fill opacity=0.60] ( 58.46, 65.77) --
	( 58.46, 53.01) --
	( 66.22, 53.01) --
	( 66.22, 65.77) --
	( 58.46, 65.77) --
	cycle;

\path[draw=drawColor,line width= 1.1pt,line join=round] ( 58.46, 58.65) -- ( 66.22, 58.65);

\path[draw=drawColor,line width= 0.6pt,line join=round] ( 75.28, 53.53) -- ( 75.28, 74.74);

\path[draw=drawColor,line width= 0.6pt,line join=round] ( 75.28, 39.36) -- ( 75.28, 29.45);
\definecolor{fillColor}{RGB}{255,176,0}

\path[draw=drawColor,line width= 0.6pt,line join=round,line cap=round,fill=fillColor,fill opacity=0.60] ( 71.40, 53.53) --
	( 71.40, 39.36) --
	( 79.17, 39.36) --
	( 79.17, 53.53) --
	( 71.40, 53.53) --
	cycle;

\path[draw=drawColor,line width= 1.1pt,line join=round] ( 71.40, 44.69) -- ( 79.17, 44.69);

\path[draw=drawColor,line width= 0.6pt,line join=round] ( 88.23, 74.79) -- ( 88.23,134.77);

\path[draw=drawColor,line width= 0.6pt,line join=round] ( 88.23, 58.61) -- ( 88.23, 32.84);
\definecolor{fillColor}{RGB}{0,192,0}

\path[draw=drawColor,line width= 0.6pt,line join=round,line cap=round,fill=fillColor,fill opacity=0.60] ( 84.34, 74.79) --
	( 84.34, 58.61) --
	( 92.11, 58.61) --
	( 92.11, 74.79) --
	( 84.34, 74.79) --
	cycle;

\path[draw=drawColor,line width= 1.1pt,line join=round] ( 84.34, 64.32) -- ( 92.11, 64.32);

\path[draw=drawColor,line width= 0.6pt,line join=round] (101.17, 70.28) -- (101.17, 88.12);

\path[draw=drawColor,line width= 0.6pt,line join=round] (101.17, 53.03) -- (101.17, 36.00);
\definecolor{fillColor}{RGB}{0,192,255}

\path[draw=drawColor,line width= 0.6pt,line join=round,line cap=round,fill=fillColor,fill opacity=0.60] ( 97.29, 70.28) --
	( 97.29, 53.03) --
	(105.05, 53.03) --
	(105.05, 70.28) --
	( 97.29, 70.28) --
	cycle;

\path[draw=drawColor,line width= 1.1pt,line join=round] ( 97.29, 61.59) -- (105.05, 61.59);

\path[draw=drawColor,line width= 0.6pt,line join=round] (114.11, 67.49) -- (114.11, 87.93);

\path[draw=drawColor,line width= 0.6pt,line join=round] (114.11, 48.56) -- (114.11, 29.45);
\definecolor{fillColor}{RGB}{160,96,255}

\path[draw=drawColor,line width= 0.6pt,line join=round,line cap=round,fill=fillColor,fill opacity=0.60] (110.23, 67.49) --
	(110.23, 48.56) --
	(117.99, 48.56) --
	(117.99, 67.49) --
	(110.23, 67.49) --
	cycle;

\path[draw=drawColor,line width= 1.1pt,line join=round] (110.23, 58.04) -- (117.99, 58.04);

\path[draw=drawColor,line width= 0.6pt,line join=round] (127.05, 71.97) -- (127.05,134.77);

\path[draw=drawColor,line width= 0.6pt,line join=round] (127.05, 51.49) -- (127.05, 29.45);
\definecolor{fillColor}{RGB}{0,192,0}

\path[draw=drawColor,line width= 0.6pt,line join=round,line cap=round,fill=fillColor,fill opacity=0.60] (123.17, 71.97) --
	(123.17, 51.49) --
	(130.94, 51.49) --
	(130.94, 71.97) --
	(123.17, 71.97) --
	cycle;

\path[draw=drawColor,line width= 1.1pt,line join=round] (123.17, 61.29) -- (130.94, 61.29);

\path[draw=drawColor,line width= 0.6pt,line join=round] (140.00, 71.27) -- (140.00,101.22);

\path[draw=drawColor,line width= 0.6pt,line join=round] (140.00, 54.35) -- (140.00, 34.24);
\definecolor{fillColor}{RGB}{0,192,255}

\path[draw=drawColor,line width= 0.6pt,line join=round,line cap=round,fill=fillColor,fill opacity=0.60] (136.11, 71.27) --
	(136.11, 54.35) --
	(143.88, 54.35) --
	(143.88, 71.27) --
	(136.11, 71.27) --
	cycle;

\path[draw=drawColor,line width= 1.1pt,line join=round] (136.11, 62.50) -- (143.88, 62.50);
\definecolor{drawColor}{RGB}{0,0,0}

\path[draw=drawColor,draw opacity=0.50,line width= 0.6pt,line join=round] ( 81.76, 24.19) -- ( 81.76,140.04);

\path[draw=drawColor,draw opacity=0.50,line width= 0.6pt,line join=round] (120.58, 24.19) -- (120.58,140.04);
\end{scope}
\begin{scope}
\path[clip] (147.76, 24.19) rectangle (161.72,140.04);
\definecolor{drawColor}{gray}{0.10}

\node[text=drawColor,rotate=-90.00,anchor=base,inner sep=0pt, outer sep=0pt, scale=  0.72] at (152.26, 82.11) {robot3};
\end{scope}
\begin{scope}
\path[clip] (  0.00,  0.00) rectangle (166.22,144.54);
\definecolor{drawColor}{gray}{0.20}

\path[draw=drawColor,line width= 0.6pt,line join=round] ( 23.51, 21.94) --
	( 23.51, 24.19);

\path[draw=drawColor,line width= 0.6pt,line join=round] ( 36.46, 21.94) --
	( 36.46, 24.19);

\path[draw=drawColor,line width= 0.6pt,line join=round] ( 49.40, 21.94) --
	( 49.40, 24.19);

\path[draw=drawColor,line width= 0.6pt,line join=round] ( 62.34, 21.94) --
	( 62.34, 24.19);

\path[draw=drawColor,line width= 0.6pt,line join=round] ( 75.28, 21.94) --
	( 75.28, 24.19);

\path[draw=drawColor,line width= 0.6pt,line join=round] ( 88.23, 21.94) --
	( 88.23, 24.19);

\path[draw=drawColor,line width= 0.6pt,line join=round] (101.17, 21.94) --
	(101.17, 24.19);

\path[draw=drawColor,line width= 0.6pt,line join=round] (114.11, 21.94) --
	(114.11, 24.19);

\path[draw=drawColor,line width= 0.6pt,line join=round] (127.05, 21.94) --
	(127.05, 24.19);

\path[draw=drawColor,line width= 0.6pt,line join=round] (140.00, 21.94) --
	(140.00, 24.19);
\end{scope}
\begin{scope}
\path[clip] (  0.00,  0.00) rectangle (166.22,144.54);
\definecolor{drawColor}{gray}{0.30}

\node[text=drawColor,rotate= 90.00,anchor=base,inner sep=0pt, outer sep=0pt, scale=  0.72] at ( 28.47, 12.32) {0.05};

\node[text=drawColor,rotate= 90.00,anchor=base,inner sep=0pt, outer sep=0pt, scale=  0.72] at ( 41.42, 12.32) {0.1};

\node[text=drawColor,rotate= 90.00,anchor=base,inner sep=0pt, outer sep=0pt, scale=  0.72] at ( 54.36, 12.32) {0.2};

\node[text=drawColor,rotate= 90.00,anchor=base,inner sep=0pt, outer sep=0pt, scale=  0.72] at ( 67.30, 12.32) {0.4};

\node[text=drawColor,rotate= 90.00,anchor=base,inner sep=0pt, outer sep=0pt, scale=  0.72] at ( 80.24, 12.32) {0.8};

\node[text=drawColor,rotate= 90.00,anchor=base,inner sep=0pt, outer sep=0pt, scale=  0.72] at ( 93.19, 12.32) {5$\times$5};

\node[text=drawColor,rotate= 90.00,anchor=base,inner sep=0pt, outer sep=0pt, scale=  0.72] at (106.13, 12.32) {7$\times$7};

\node[text=drawColor,rotate= 90.00,anchor=base,inner sep=0pt, outer sep=0pt, scale=  0.72] at (119.07, 12.32) {9$\times$9};

\node[text=drawColor,rotate= 90.00,anchor=base,inner sep=0pt, outer sep=0pt, scale=  0.72] at (132.01, 12.32) {all};

\node[text=drawColor,rotate= 90.00,anchor=base,inner sep=0pt, outer sep=0pt, scale=  0.72] at (144.96, 12.32) {some};
\end{scope}
\begin{scope}
\path[clip] (  0.00,  0.00) rectangle (166.22,144.54);
\definecolor{drawColor}{gray}{0.30}

\node[text=drawColor, anchor=base,inner sep=0pt, outer sep=0pt, scale=  0.72] at (50.89, 134.77) { Magnitude };
\node[text=drawColor, anchor=base,inner sep=0pt, outer sep=0pt, scale=  0.72] at (101.87, 134.77) { Resolution };
\node[text=drawColor, anchor=base,inner sep=0pt, outer sep=0pt, scale=  0.72] at (133.6, 134.77) { Type };

\node[text=drawColor,anchor=base east,inner sep=0pt, outer sep=0pt, scale=  0.72] at ( 11.70, 47.20) {4};

\node[text=drawColor,anchor=base east,inner sep=0pt, outer sep=0pt, scale=  0.72] at ( 11.70, 76.03) {6};

\node[text=drawColor,anchor=base east,inner sep=0pt, outer sep=0pt, scale=  0.72] at ( 11.70,104.85) {8};

\node[text=drawColor,anchor=base east,inner sep=0pt, outer sep=0pt, scale=  0.72] at ( 11.70,133.68) {10};
\end{scope}
\begin{scope}
\path[clip] (  0.00,  0.00) rectangle (166.22,144.54);
\definecolor{drawColor}{gray}{0.20}

\path[draw=drawColor,line width= 0.6pt,line join=round] ( 13.50, 49.68) --
	( 15.75, 49.68);

\path[draw=drawColor,line width= 0.6pt,line join=round] ( 13.50, 78.50) --
	( 15.75, 78.50);

\path[draw=drawColor,line width= 0.6pt,line join=round] ( 13.50,107.33) --
	( 15.75,107.33);

\path[draw=drawColor,line width= 0.6pt,line join=round] ( 13.50,136.16) --
	( 15.75,136.16);
\end{scope}
\end{tikzpicture}
\\
Coverage
&
Reliability
&
Precision
\end{tabular}
\caption{Box plots of the three objectives for robot3 grouped by parameter. 
The vertical axes measure the performance measure given by the subtitle. The horizontal ticks name the parameter values: the $\sigma$, map resolutions and mutation type parameters are separated by vertical lines.  
}
\label{fig:boxes}
\end{figure*}
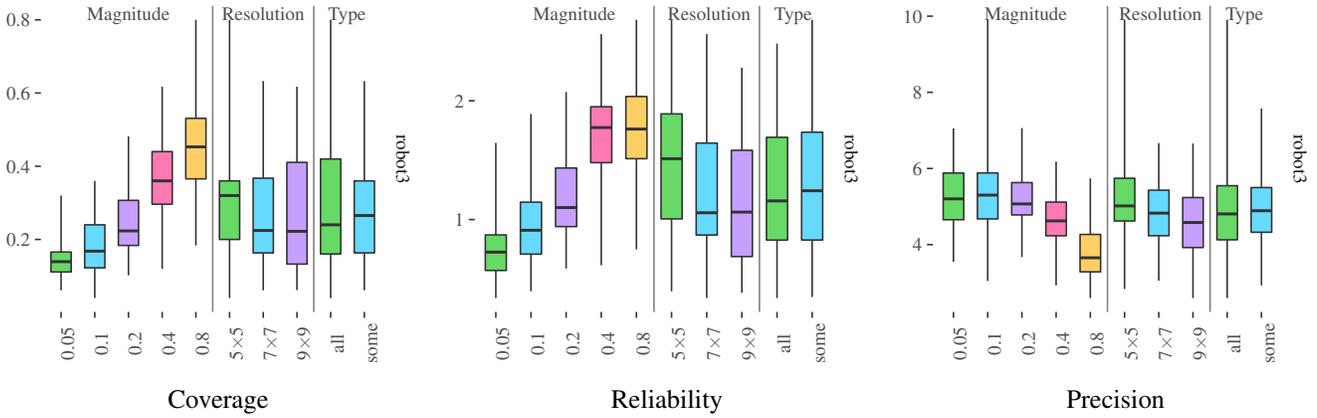

\subsection{Pareto Domination}\label{method:pareto}
Considering the performance measures defined in \autoref{sec:perfmeas} from the perspective of multi-objective optimization, we have three objectives we want to optimize. Rather than analyzing algorithm performance with regards to the objectives individually, we are interested in performance across all objectives. Pareto domination provides a way to test this. 

An objective vector $\mathbf x$ is said to dominate another vector $\mathbf y$ if all individual objectives are at least as good, and at least one objective is better:
\[
x \succ_\mathsf{P} y \ \ \equiv\ \ 
\forall i\ x_i \unrhd y_i \ \ \land\ \ 
\exists i\ x_i \rhd y_i 
\]

As suggested by \cite{Neumann2015}, we can replace the comparison operators in the equation defining Cliff's delta (\ref{eq:d}) with Pareto domination, resulting in a measure of what degree one sample dominates the other:
\begin{equation}
	d_{ab} = P(a \succ_\mathsf{P} b) - P(a \prec_\mathsf{P} b)
\end{equation} 

Like the ordinary Cliff's delta, $\delta_{ab}$, $d_{ab}$ can be used to compare distributions to decide which is better. Note, however, that the WMW U statistic and test cannot be generalized in the same manner, so it cannot be used to calculate the Pareto d statistic. This is because Pareto dominance is not able to produce a simple ranking of a sample.

Note that given the way \precision, \coverage and \reliability is defined here, reliability is redundant when considering Pareto domination: consider if, for some maps $m$ and $n$, $\opp Gm < \opp Gn$, thus implying $m \nsucc_\mathsf{P} n$. Expanding this as products of $\op P$ and $\op C$ we get
\[ \opp Pm \opp Cm < \opp Pn \opp Cn \]
However, since both $\op P$ and $\op C$ are always non-negative, this would imply $\opp Pm < \opp Pn \lor \opp Cm < \opp Cn$, and then we would already have $m \nsucc_\mathsf{P} n$ based on \precision and \coverage alone.

\section{Results}

\autoref{fig:boxes} shows the conventional box plots for \coverage, \reliability and \precision after the last evaluation for robot3.
The plots indicate that mutation magnitude has the largest effect on all three measures. \Coverage and \reliability increases with mutation magnitude, while high mutation magnitudes seem to have a negative effect on \precision. There also seems to be a small negative effect on all three measures from increasing map resolution. There is little or no difference between the mutation types on any of the measures.
Corresponding plots for robot2, robot4, and robot5 are omitted, but mostly show the same main trends, with only robot2 deviating notably from the patterns, and those differences are mainly visible in a change of scales.


\autoref{fig:bagplots} shows coverage-precision plots of the results by robot and parameter. 
In the plots, mutation magnitude has the most visible effect. The traces spread out from a common origin into different niches in order from least magnitude, least \coverage, to largest magnitude, most \coverage. Magnitudes 0.2, 0.4 and 0.8 forms a clear Pareto front, while 0.1 and 0.05 are behind 0.2 on average on most robots. There is considerable overlap in the half-dominated areas, but a clear distinction between the areas of the extreme parameter values for all robots but robot2. In the plots for the other two parameters there are not any such clear patterns. The shaded areas for mutation type almost completely overlap, with what is perhaps a slight advantage to some-soft on average. Of the three map resolutions, $5\times5$ is considerably better on average, but has especially poor \coverage on robot2.

Of the four morphologies, robot2 stands out by having much higher \precision and lower \coverage than the others, along with correspondingly higher \precision variance and lower \coverage variance. The others generally have about the same shape and average development; however, robot3 appears to achieve better \coverage with higher mutation sizes, while robot4 has bigger variation in \precision.

\begin{table}
\caption{Pareto-based Cliff's delta effect sizes for each pair of parameter values }
\centering\small
\begin{tabular}{cc}
\makecell[c]{
Map resolutions$^m$ \\
%
\begin{tabular}{ccc}
\hline
 \diagbox[height=1.5em]{$a$}{$b$} & 5 & 7 \\ 
  \hline
  7 & \hspace{-6pt}$\begin{array}{c}\mathbf{-0.172} \\ \pm0.051\end{array}\hspace{-6pt}$ &  \\ 
  9 & \hspace{-6pt}$\begin{array}{c}\mathbf{-0.264} \\ \pm0.055\end{array}\hspace{-6pt}$ & \hspace{-6pt}$\begin{array}{c}\mathbf{-0.147} \\ \pm0.057\end{array}\hspace{-6pt}$ \\ 
   \hline
\end{tabular}
 \\
$5 \succ 7 \succ 9$ 
} & 
\makecell[c]{
Mutation types$^{mr}$ \\
%
\begin{tabular}{cc}
\hline
 \diagbox[height=1.5em]{$a$}{$b$} & all \\ 
  \hline
  some & \hspace{-6pt}$\begin{array}{c}-0.046 \\ \pm0.056\end{array}\hspace{-6pt}$ \\ 
   \hline
\end{tabular}
 
}
\end{tabular}
\\
\vspace{6pt}
Mutation magnitudes$^r$ \\
%
\begin{tabular}{ccccc}
 \hline
 \diagbox[height=1.5em]{$a$}{$b$} & 0.05 & 0.1 & 0.2 & 0.4 \\ 
  \hline
  0.1 & \hspace{-6pt}$\begin{array}{c}\mathbf{+0.236} \\ \pm0.078\end{array}\hspace{-6pt}$ &  &  &  \\ 
  0.2 & \hspace{-6pt}$\begin{array}{c}\mathbf{+0.393} \\ \pm0.069\end{array}\hspace{-6pt}$ & \hspace{-6pt}$\begin{array}{c}\mathbf{+0.180} \\ \pm0.072\end{array}\hspace{-6pt}$ &  &  \\ 
  0.4 & \hspace{-6pt}$\begin{array}{c}\mathbf{+0.337} \\ \pm0.066\end{array}\hspace{-6pt}$ & \hspace{-6pt}$\begin{array}{c}\mathbf{+0.260} \\ \pm0.067\end{array}\hspace{-6pt}$ & \hspace{-6pt}$\begin{array}{c}\mathbf{+0.130} \\ \pm0.071\end{array}\hspace{-6pt}$ &  \\ 
  0.8 & \hspace{-6pt}$\begin{array}{c}\mathbf{+0.153} \\ \pm0.058\end{array}\hspace{-6pt}$ & \hspace{-6pt}$\begin{array}{c}\mathbf{+0.089} \\ \pm0.058\end{array}\hspace{-6pt}$ & \hspace{-6pt}$\begin{array}{c}-0.006 \\ \pm0.061\end{array}\hspace{-6pt}$ & \hspace{-6pt}$\begin{array}{c}\mathbf{-0.131} \\ \pm0.077\end{array}\hspace{-6pt}$ \\ 
   \hline
\end{tabular}
 \\
$0.4 \succ \left\lbrace 0.8, 0.2 \right\rbrace \succ 0.1 \succ 0.05$ \\
\scriptsize
\explanation{
Effect sizes statistically significantly different from zero are in bold. When the effect sizes implies a (partial) ordering, it is shown beneath the table. 
\setlength\tabcolsep{2pt}
\begin{tabular}{rl}
 $^m$&\textrm{Grouped by mutation magnitude} \\
 $^r$&\textrm{Grouped by map resolution} 
\end{tabular}
  }

\label{tbl:effectsizes}
\end{table}

\begin{figure}
\input{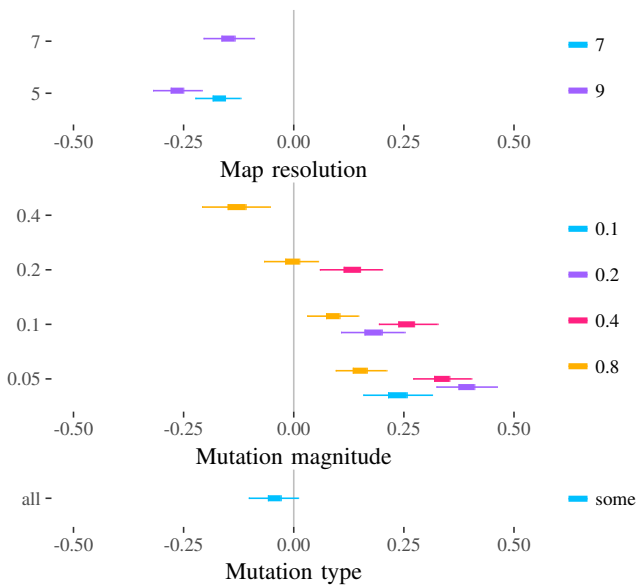}
\vspace{-9pt}
\caption{Graphical representation of the effect sizes. 
Each row in the figure corresponds to a column in \autoref{tbl:effectsizes}. The colored lines indicate the confidence interval for $d_{ab}$ with $a$ given by the color and $b$ given by the row. The thin lines show the 99\% confidence interval, while the thick lines represent the central 50\%. If the thin line does not cross the zero line then one of the groups are statistically significantly better. }
\label{fig:effectsizes}
\end{figure}

\begin{figure*}
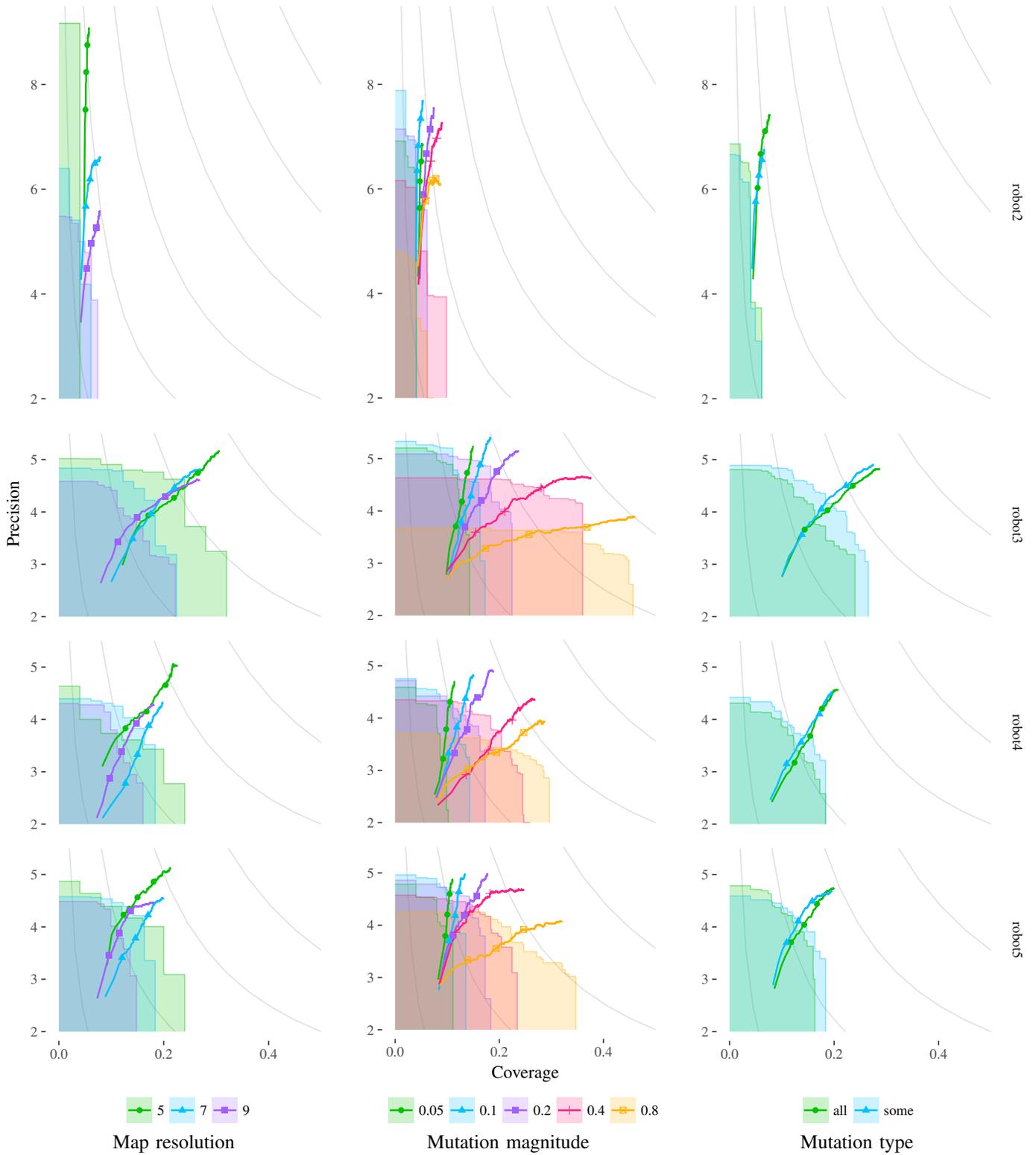

\setlength\tabcolsep{0pt}
\setlength{\fboxsep}{0pt}
\begin{tabular}{ccc}
\hspace{-0pt}{\input{bags-mapsize.tex}}
&
\hspace{-0pt}{\input{bags-sigma.tex}}
&
\hspace{-0pt}{\input{bags-mutation.tex}}
\\
Map resolution 
&
Mutation magnitude
&
Mutation type 
\end{tabular}
{
}

{
}
\caption{Precision and coverage grouped by the different variables. 
The curves trace the mean of each group as the evaluation count increases, with points marking the 350th, 1250th and 5000th evaluation. The shaded areas are the areas dominated by at least half of the runs after the last evaluation. The light gray curves are isocurves for reliability. }
\label{fig:bagplots}
\end{figure*}

\autoref{tbl:effectsizes} and \autoref{fig:effectsizes} shows the Pareto-based Cliff's deltas for all pairs of values of each parameter. The confidence intervals are 99\% confidence intervals computed by bootstrapping. Because checking if a $1-\alpha$ confidence interval overlaps with some $x$ is equivalent to a hypothesis test with threshold $\alpha$ on whether the true value is $x$ \cite{Greenland2016}, we test for statistically significant differences ($\alpha = 0.01$) between groups by checking if the reported confidence intervals overlap with zero.

For the map resolutions, the effect sizes imply a strong ordering preferring smaller map resolutions: $5\times5$ is likely to produce a better result than both $7\times7$ and $9\times9$, and $7\times7$ is likely to produce a better result than $9\times 9$. 

For the mutation magnitudes, the effect sizes form a partial, but almost complete ordering: $0.05$ is worse than all other values. $0.1$ is worse than all values but $0.05$. $0.2$ and $0.8$ are equally good, both better than $0.05$ and $0.1$ and worse than $0.4$. 

For the mutation types, the effect size is quite small, and we can not rule out that there is no difference at all between the two methods.

\section{Discussion}
From the conventional plot in \autoref{fig:boxes} one can identify that for robot3 lower map resolutions lead to higher \precision while \coverage remains about the same. The different $\sigma$ values lead to different trade-offs in terms of \precision and \coverage, while the two mutation types are hard to differentiate. These observations are all confirmed by \autoref{fig:bagplots} to hold for robot4 and robot5 as well, and to a certain degree also for robot2. These plots also indicate that $\sigma = 0.4$, which for most robots has the largest half-dominating area and the largest part not covered by other areas, might be the best overall choice. The areas for mutation type also indicate that mutating all parameters may have a slight advantage over mutating only some. 
The reported effect sizes in \autoref{tbl:effectsizes} and \autoref{fig:effectsizes} captures all these observations, quantified and generalized across the four robots.

Because of the large number of different parameter values in the experiment, showing the box plots for all four robots would require the same amount of space as \autoref{fig:bagplots} does, but to us it is clear that \autoref{fig:bagplots} gives a significantly clearer view of the performance trade-offs of a set of parametrizations this large.
In addition, plotting along the coverage-precision dimensions allows us to trace the search progress with systematic marks.
As an obervation from this, there is a tendency for the exponentially placed marks to be evenly spaced in the plot, indicating a trend of exponential convergence.

From the mutation magnitude plot for robot3 in \autoref{fig:bagplots}, we can see that 0.4 and 0.8 are one the same isocurve, which results in in similar reliability values. Still, the coverage-precision plot highlights the difference in these parametrizations, and makes it easier to select based on the application. To achieve the same insights with the box plots, one would need to cross-reference at least two of them.





\section{Conclusion}
We have presented a method for multi-objective performance analysis of the \ME algorithm, based on ordinal effect sizes and Pareto dominance.
Since the method uses ordinal effect size, it allows us to draw general conclusions on the performance of various parameter values across groups of different scaling, such as varying robot morphologies.

Through a thorough case study we demonstrated that this approach allowed us to better discern performance in trade-off scenarios as seen when varying the $\sigma$ parameter.
At the same time, it reproduced the conclusions where the traditional analysis has proven robust. 

We expect the method to be useful for other \ME practitioners, both for analyzing new algorithmic features as well as for tuning performance to specific applications.
We also expect that the method could find application beyond \ME, in particular it should be applicable to other algorithms within the Quality Diversity domain.


\footnotesize
\bibliographystyle{IEEEtran}
\bibliography{../../bibtex/kyrrehg}

\end{document}